\documentclass[runningheads]{llncs}
\usepackage{graphicx}
 
\usepackage{subcaption}

\usepackage{eccv}
\usepackage{fontawesome5}

\usepackage{booktabs}
\usepackage{tabularx}
\usepackage[table]{xcolor}
\usepackage{ragged2e}
\newcolumntype{L}[1]{>{\RaggedRight\arraybackslash}p{#1}}

\usepackage{eccvabbrv}
\usepackage{multirow}

\usepackage{booktabs}
\usepackage{tabularx}      
\usepackage{threeparttable}
\usepackage{makecell}      
\usepackage[table]{xcolor} 
\usepackage{array}         
\usepackage{colortbl}      
\usepackage{adjustbox}

\usepackage{xcolor}
\usepackage{pifont}
\usepackage{subcaption}
\usepackage{float}
\definecolor{TickGreen}{HTML}{1B8E3E}
\definecolor{OpticalBlue}{HTML}{E9F2FF}
\definecolor{SARGreen}{HTML}{E9FBE9}
\definecolor{GISOrange}{HTML}{FFF3E0}
\definecolor{IndexPurple}{HTML}{F3E9FF}

\definecolor{NeonGreen}{RGB}{57,255,20}

\newcommand{\cmark}{\textcolor{TickGreen}{\ding{51}}}
\newcommand{\xmark}{\textcolor{black}{\ding{55}}}

\usepackage[accsupp]{axessibility}  

\usepackage{hyperref}

\usepackage{orcidlink}

\begin{document}


\title{OpenEarthAgent\protect\raisebox{-0.15ex}{\includegraphics[height=1.5em]{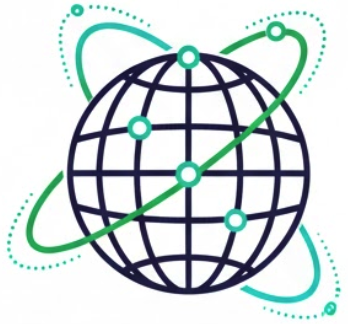}}: A Unified Framework for Tool-Augmented Geospatial Agents}
\titlerunning{OpenEarthAgent} 
\author{\small
Akashah Shabbir\inst{1}\thanks{Equal contribution.} \and
Muhammad Umer Sheikh\inst{1}\protect\footnotemark[1] \and
Muhammad Akhtar Munir\inst{1} \and \\
Hiyam Debary\inst{2} \and
Mustansar Fiaz\inst{2} \and
Muhammad Zaigham Zaheer\inst{1} \and
Paolo Fraccaro\inst{2} \and \\
Fahad Shahbaz Khan\inst{1,3} \and
Muhammad Haris Khan\inst{1} \and
Xiao Xiang Zhu\inst{4} \and \\
Salman Khan\inst{1,5}
}

\authorrunning{A. Shabbir et al.}

\institute{
Mohamed bin Zayed University of AI, United Arab Emirates \and IBM Research, Europe \and Linköping University, Sweden \and Technical University Munich, Germany \and Australian National University, Australia 
\small\faGlobe \url{https://github.com/mbzuai-oryx/OpenEarthAgent}
}

\maketitle
\vspace{-10pt}
\begin{figure}[!h]
  \centering
  \includegraphics[width=1\linewidth]{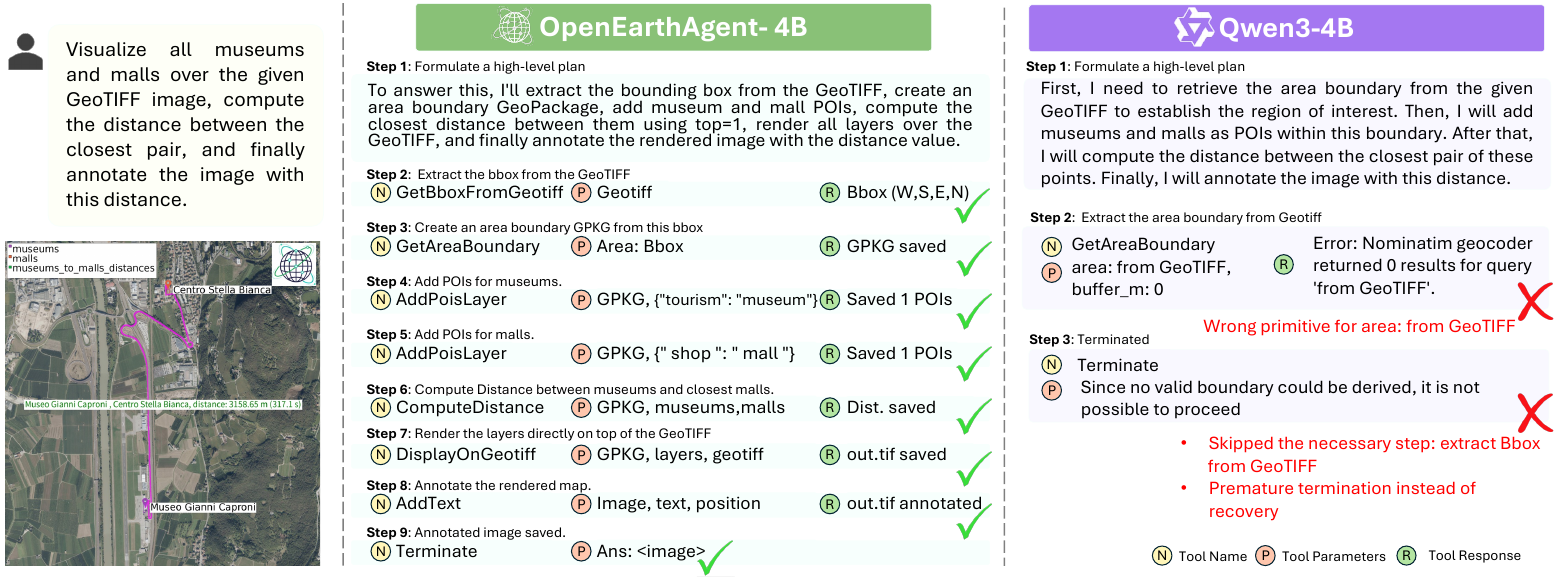}
  \caption{Comparison of OpenEarthAgent-4B and Qwen3-4B on a complex GIS reasoning task. 
  OpenEarthAgent correctly sequences tool calls with proper dependencies and feedback, while Qwen3 fails due to misordered tool usage and inconsistent reasoning.
  } 
  \label{fig:teaser}
\end{figure}
\vspace{-20pt}

\begin{abstract}
Recent progress in multimodal reasoning has enabled agents that interpret imagery, connect it with language, and execute structured analytical tasks.
Extending these capabilities to remote sensing remains challenging, as models must reason over spatial scale, geographic structures, and multispectral indices while maintaining coherent multi-step logic.
To address this gap, we introduce \textit{OpenEarthAgent}, a unified framework for tool-augmented geospatial reasoning trained on satellite imagery, natural-language queries, and structured reasoning traces.
Beyond serving as a benchmark, OpenEarthAgent establishes a cohesive agentic architecture built around a unified executable tool registry and trajectory-based policy learning.
The framework standardizes heterogeneous visual, spectral, GIS, and georeferenced raster operations under a consistent callable schema, enabling modular orchestration and deterministic execution.
Training is performed via supervised fine-tuning on structured reasoning trajectories with deterministic replay validation to ensure executability and spatial correctness.
The accompanying corpus comprises 14,538 training and 1,169 evaluation instances with over 107K reasoning steps, spanning urban, environmental, disaster, and infrastructure domains and incorporating GIS operations alongside index analyses such as NDVI, NBR, and NDBI.
Grounded in explicit reasoning traces, the learned agent demonstrates structured reasoning, stable spatial understanding, and interpretable tool-driven behavior across diverse EO scenarios.
We report consistent improvements over a strong baseline and competitive performance against recent open and closed-source models.
\end{abstract}

\section{Introduction}
\label{sec:intro}

The evolution of visual representation learning has transitioned from static perception to interactive multimodal reasoning.
Early single-shot frameworks such as DINO \cite{caron2021emerging} and MAE \cite{he2022masked} laid the foundation for large-scale visual understanding through self-supervised objectives, learning strong image encoders without explicit language-based reasoning or interaction.
Inspired by these successes, the remote sensing community extended this paradigm to earth observation, where models such as Prithvi \cite{jakubik2023foundation}, Copernicus-FM \cite{wang2025towards}, Galileo \cite{tseng2025galileo}, Panopticon \cite{waldmann2025panopticon}, TerraFM \cite{danish2025terrafm}, and AnySat \cite{astruc2025anysat} scaled representation learning to global multisensor data.
These earth-scale vision models demonstrated remarkable transfer across spatial resolutions and modalities, yet their predictions remained one-shot and perception-centric, focusing on recognition rather than structured reasoning.

In parallel, large multi-modal models have progressed beyond perception.
Architectures such as BLIP-2 \cite{li2023blip}, InstructBLIP \cite{dai2023instructblip}, LLaVA-OneVision \cite{li2024llava}, and Kosmos-2 \cite{peng2023kosmos} extended language models into the visual domain, enabling grounded image understanding.
Building on this trajectory, remote-sensing VLMs emerged to adapt multimodal reasoning to geospatial data.
Early efforts such as RemoteCLIP \cite{liu2024remoteclip}, SkySenseGPT \cite{luo2024skysensegpt}, and GeoChat \cite{kuckreja2024geochat} introduced large-scale multimodal alignment for remote sensing through paired image–text and instruction-following datasets.
Recently, EarthDial \cite{soni2025earthdial} advanced this paradigm by extending grounding across optical, SAR, thermal, and temporal modalities, enabling unified multimodal reasoning for classification, captioning, and change analysis.
Despite coupling language with perception, current remote sensing VLMs remain largely descriptive and lack explicit structured reasoning.

Subsequent reasoning-centric approaches such as ReAct \cite{yao2022react}, DeepSeek-R1 \cite{guo2025deepseek}, and VLM-R1 \cite{shen2025vlm} introduced structured planning and, in some cases, tool-driven reasoning, transforming static perception into organized multi-step task execution.
Recent developments such as OpenThinkIMG \cite{su2025openthinkimg} unified these advances through multi-step vision-tool interaction, where large vision-language models iteratively ``\textit{think with images}'' through executable reasoning trajectories.
Inspired by these advances, efforts such as ThinkGeo \cite{shabbir2025thinkgeo} and Earth-Agent \cite{feng2025earth} began exploring tool-augmented reasoning for earth observation (EO).
They demonstrate that large models can plan analytical chains but still face challenges in geospatial grounding, coordinate consistency, and physically verifiable outputs.
These limitations highlight the need for geographically grounded agents that integrate perception with explicit and interpretable reasoning.

Building on this motivation, we construct a comprehensive agentic corpus and training framework that integrates GIS layers, index-driven computations, and multisensor imagery (optical and SAR) modalities to support a wide spectrum of EO reasoning tasks.
Our dataset comprises training image-query reasoning-traces, along with a held-out evaluation set designed to benchmark model consistency and reasoning performance.
The data span diverse contexts, including urban infrastructure, environmental monitoring, disaster assessment, land-use mapping, and transportation analysis, and incorporate GIS operations (e.g., distance, area, zonal statistics) alongside index-based analyses such as NDVI, NBR, and NDBI.
Each sample provides an explicit reasoning trajectory linking multiple tool calls, intermediate states, and outcomes, enabling agents to learn structured, interpretable workflows rather than static predictions.

Central to this design is a unified tool registry that abstracts heterogeneous geospatial operators through a standardized executable schema, enabling consistent orchestration across perception, GIS computation, spectral analysis, and georeferenced raster reasoning.
Unlike prior EO benchmarks that focus primarily on data scale, our contribution lies in coupling the dataset with an execution engine and trajectory-based policy alignment.
This integration ensures that reasoning steps are not only generated but deterministically validated for spatial and logical consistency.
To assess generalization and reasoning fidelity, we benchmark a broad suite of large and reasoning models, including the GPT family \cite{OpenAI2025GPT}, Qwen \cite{yang2025qwen3}, InternLM \cite{cai2024internlm2}, and other recent multimodal LLMs.
Our dataset is constructed through a unified data-gathering pipeline, where each modality undergoes question synthesis and automated validation to ensure reasoning diversity, spatial grounding, and cross-source consistency.
This large-scale, tool-augmented corpus bridges the divide between GIS analysis and remote-sensing perception, providing a unified setting for evaluating structured geospatial reasoning.
This work introduces an agentic framework for remote sensing that integrates tool-based reasoning within a unified training and evaluation setup. 

Specifically, our key contributions are:
\begin{itemize}
    \item \textbf{Unified data-construction pipeline:} A systematic process integrating multisensor imagery (RGB, SAR), GIS layers, and index-based computations through question synthesis and automated validation.
    \item \textbf{Agentic reasoning framework with deterministic validation:} A unified tool registry and a trajectory-based supervised alignment mechanism that enforces consistent tool syntax, executable reasoning chains, and spatially grounded decision-making (Fig.~\ref{fig:teaser}).
    \item \textbf{Comprehensive multimodal corpus:} 14,538 training instances and 1,169 evaluation tasks establishing a benchmark for studying spatial reasoning, grounding, and interpretability across reasoning and non-reasoning models.
\end{itemize}

\section{Related Work}

Remote sensing has progressed from task-specific models to large-scale foundation and vision-language models trained on global, multimodal data.  
Early self-supervised methods \cite{manas2021seasonal, cong2022satmae} established transferable pretraining strategies on Sentinel archives.  
Subsequent foundation models include Prithvi-v2 \cite{szwarcman2024prithvi} with spatiotemporal transformers on HLS, Copernicus-FM \cite{wang2025towards} with metadata-aware hypernetworks for Sentinel, Panopticon \cite{waldmann2025panopticon} and AnySat \cite{astruc2025anysat} with spectral- and resolution-adaptive embeddings, and CROMA \cite{fuller2023croma} fusing radar and optical contrastive learning. Galileo \cite{tseng2025galileo} captures global-to-local context, demonstrating scalable pretraining across diverse sensors and geographies.  
On the multimodal side, GeoChat \cite{kuckreja2024geochat} enabled language grounding to satellite imagery, and EarthDial \cite{soni2025earthdial} extended this paradigm to multi-resolution optical, SAR, thermal, and temporal modalities across diverse tasks.  
Despite their scalability and representational strength, these models remain largely single-pass encoders; they lack structured reasoning, tool orchestration, and verifiable intermediate analysis.

Alongside foundation models, agentic systems have emerged that couple reasoning with external tools and feedback-driven control.  
Early frameworks such as ReAct \cite{yao2022react} and Voyager \cite{wang2023voyager} demonstrated the transition from static inference to autonomous, goal-directed reasoning.  
Subsequent systems, including WebAgent \cite{wei2025webagent}, VisTA \cite{huang2025visualtoolagent}, and DeepEyes \cite{zheng2025deepeyes}, introduced modular architectures capable of selecting and sequencing tools under guided supervision, improving reasoning depth and execution consistency.  
OpenThinkIMG \cite{su2025openthinkimg} unified these ideas through large-scale visual--tool integration with standardized APIs, while OctoTools \cite{lu2025octotools} emphasized modular, verifiable execution via structured tool interfaces.  
Collectively, these systems established key principles: looped reasoning, standardized tool I/O, and trajectory-based learning, yet remain geospatially naive, lacking coordinate awareness, scale handling, and domain-specific spatial verification essential for earth observation analysis.

An emerging wave of EO research is now bringing these agentic principles into geospatial contexts.  
ThinkGeo \cite{shabbir2025thinkgeo} frames EO QA as tool-augmented reasoning, benchmarking ReAct-style chains and exposing persistent weaknesses in coordinate consistency, spatial grounding, and multi-step planning.  
Earth-Agent \cite{feng2025earth} broadens the tool ecosystem to include spectral products and standardized interfaces but still relies on largely predefined workflows, limiting physically verifiable GIS and index-based reasoning.  
Geo-OLM \cite{stamoulis2025geo} explores plan-tool-verify prompting for compact models, improving efficiency but depending primarily on prompt-level heuristics rather than learned adaptive policies.  

Building on this gap, we introduce an EO-native agentic stack that unifies dataset construction, supervised reasoning alignment, and evaluation under a physically grounded framework.  
OpenEarthAgent enables broad orchestration of GIS, index-based, optical, and SAR operations through structured instruction-tool pairs; trains on detailed reasoning traces with a dedicated evaluation set; and supports interpretable, verifiable, multi-step reasoning across diverse geospatial conditions.

\section{OpenEarthAgent}
\subsection{Dataset Curation}
We curate a large-scale remote-sensing dataset linking imagery, queries, and tool-based reasoning traces to enable interpretable, multi-step geospatial analysis.

\noindent\textbf{Overview and Motivation:}
The dataset presented in this work serves as a foundation for training and evaluating tool-augmented geospatial reasoning agents. 
While recent EO datasets primarily focus on visual classification or retrieval, they lack explicit reasoning structure and tool-level traceability. 
Our dataset bridges this gap by integrating imagery, natural-language queries, reasoning traces, and tool executions in a unified schema. 
It enables models to perform planning, verification, and computation beyond pattern recognition. 
A comparison against existing agent frameworks is provided in Table~\ref{tab:agent_comparison}, highlighting differences in modality coverage, tool integration, reasoning depth, and supervision design.
Consisting of 14,538 training samples and 1,169 held-out test samples for benchmarking, covering diverse geographic regions, sensor modalities, and reasoning tasks, it enables interpretable multi-step learning and evaluation.

\begin{table}[t]
\centering
\caption{\textbf{Comparison of geospatial/EO agent frameworks.}
Abbrev.: RS=remote sensing; MM=multimodal inputs; Tools=executable tool/function calls;
MS=multi-step reasoning;
GIS=GIS/vector-raster geospatial;
Spec=spectral (e.g., indices, multispectral); CD=change detection; 
RData=reasoning-supervision data for training.
}
\scriptsize
\renewcommand{\arraystretch}{1.05}
\setlength{\tabcolsep}{4pt}

\rowcolors{3}{gray!5}{white}

\begin{tabularx}{\textwidth}{l *{7}{>{\centering\arraybackslash}X}}
\toprule
\rowcolor{OpticalBlue!60}
\textbf{Method} &
\textbf{RS} &
\textbf{MM} &
\textbf{MS} &
\textbf{GIS} &
\textbf{Spec} &
\textbf{CD} &
\textbf{RData} 
\\
\midrule

ReAct~\cite{yao2022react} &
\xmark & \xmark & \cmark & \xmark & \xmark & \xmark & \xmark \\

OpenThinkIMG~\cite{su2025openthinkimg} &
\xmark & \xmark & \cmark & \xmark & \xmark & \xmark & \cmark \\

ThinkGeo~\cite{shabbir2025thinkgeo} &
\cmark & \cmark & \cmark & \xmark & \xmark & \cmark & \xmark \\

Earth-Agent~\cite{feng2025earth} &
\cmark & \cmark & \cmark & \cmark & \cmark & \xmark & \xmark \\

Geo-OLM~\cite{stamoulis2025geo} &
\cmark & \xmark & \cmark & \xmark & \xmark & \xmark & \xmark \\

RS-Agent~\cite{xu2024rs} &
\cmark & \xmark & \xmark & \xmark & \xmark & \xmark & \xmark \\

RS-ChatGPT~\cite{guo2024remote} &
\cmark & \xmark & \xmark & \xmark & \xmark & \xmark & \xmark \\

\midrule
\rowcolor{OpticalBlue!60}
\textbf{OpenEarthAgent (Ours)} &
\cmark & \cmark & \cmark & \cmark & \cmark & \cmark & \cmark \\
\bottomrule
\end{tabularx}
\label{tab:agent_comparison}
\end{table}

\begin{table}[h]
\centering
\caption{Comprehensive overview of data sources used for dataset construction, across optical, SAR, GIS, and multispectral domains. These sources provide complementary resolutions and annotations spanning urban and transport monitoring to environmental and disaster assessment, forming a foundation for multi-sensor, tool-based geospatial reasoning. GSD: ground sample distance, B-Box: bounding box, Seg: segmentation.
}
\label{tab:datasets-tasks}
\scriptsize
\renewcommand{\arraystretch}{0.9}
\setlength{\tabcolsep}{3pt}
\begin{threeparttable}
\begin{minipage}{\textwidth}
\centering
\begin{tabularx}{\textwidth}{
  p{2.8cm}
  p{2.9cm}
  p{1.2cm}
  p{0.8cm}
  >{\scriptsize\raggedright\arraybackslash}X
}
\toprule
\textbf{Source} & \textbf{Annotation Type} & \makecell{\textbf{Sensor}\\\textbf{(m/px)}} & \textbf{Year} & \textbf{Applications / Tasks} \\
\midrule

\rowcolor{OpticalBlue!60}
\multicolumn{5}{l}{\textsc{Optical Datasets}} \\

DIOR \cite{li2020object} & B-Box, Category & 0.5--30 & 2024 & Transport \& Aviation Monitoring; Infrastructure Mapping \\
DOTA \cite{xia2018dota} & GSD, B-Box, Category & 0.1--1 & 2021 & Transport Surveillance; Infrastructure Assessment \\
NWPU-VHR-10 \cite{wang2018multiscale} & B-Box, Category & 0.5--2 & 2023 & Aviation; Transport; Infrastructure \\
UCAS-AOD \cite{zhu2015orientation} & B-Box, Category & 0.5--2 & 2015 & Aviation; Transport \\
AID \cite{xia2017aid} & B-Box, Category & 0.2--2 & 2017 & Urban Planning; Industrial Site Detection \\
iSAID \cite{waqas2019isaid} & GSD, B-Box, Seg. Map, Pix. Count & 0.1--1 & 2019 & Segmentation; Transport Monitoring \\
xBD \cite{gupta2019xbd} & GSD, B-Box, Category, Pix. Count & 1--3.5 & 2019 & Disaster Assessment; Change Detection \\
FloodNet \cite{rahnemoonfar2021floodnet} & GSD, B-Box, Category, Seg. Map, Pix. Count & 0.015--0.02 & 2020 & Flood Monitoring; Damage Estimation \\
Global-Dumpsite \cite{sun2023revealing} & B-Box, Category & 0.3--0.8 & 2023 & Environmental Monitoring; Waste Localization \\
\midrule

\rowcolor{SARGreen!60}
\multicolumn{5}{l}{\textsc{SAR Datasets}} \\

SSDD \cite{zhang2021sar} & B-Box, Category & 1--15 & 2021 & Ship Detection; Maritime Transport \\
SADD \cite{zhang2022sefepnet} & B-Box, Category & 0.5--3 & 2022 & Aviation; Vessel Identification \\
SIVED \cite{lin2023sived} & B-Box, Category & 0.1--0.3 & 2023 & Vehicle Detection; Transport Monitoring \\
\midrule

\rowcolor{GISOrange!60}
\multicolumn{5}{l}{\textsc{GIS-based Sources}} \\

OpenStreetMap \cite{OpenStreetMap} & POIs (Points of Interest) & N/A & 2017- & Urban Mapping; Infrastructure Analysis \\
\midrule

\rowcolor{IndexPurple!60}
\multicolumn{5}{l}{\textsc{Indexed / Multispectral Sources}} \\

GoogleEarthEngine \cite{GoogleEarthEngine} & Index Layers & 10--30 & 2017- & Urban Planning; Change Detection; Environmental Monitoring \\

\bottomrule
\end{tabularx}
\end{minipage}
\end{threeparttable}
\end{table}

\noindent\textbf{Data Sources and Composition:}
The dataset integrates heterogeneous imagery and metadata from open-access EO repositories, encompassing optical, SAR, GIS, and multispectral domains. 
Primary sources include high-resolution benchmarks such as DOTA \cite{xia2018dota}, DIOR \cite{li2020object}, xBD \cite{gupta2019xbd}, AID \cite{xia2017aid}, NWPU-VHR-10 \cite{wang2018multiscale}, FloodNet \cite{rahnemoonfar2021floodnet}, and Global-Dumpsite \cite{sun2023revealing}, complemented by Sentinel collections and Google Earth Engine archives (see Table~\ref{tab:datasets-tasks} for details). 
Each dataset offers distinct spatial resolutions and annotation formats (bounding boxes, segmentation masks, category labels), providing both fine-grained object understanding and large-scale semantic context.
The collection spans seven thematic domains: \textit{urban planning, disaster assessment, environmental monitoring, transportation, aviation, recreation, and industrial infrastructure}. 
Multispectral and indexed layers (e.g., NDVI, NBR, NDBI) are derived through Google Earth Engine to complement optical and SAR imagery with physically meaningful environmental indicators.
Where required for spatial reasoning tasks, samples are associated with geospatial metadata such as coordinates, projection reference, and ground sampling distance (GSD), enabling verifiable geospatial analysis.

\begin{figure*}[t]
  \centering
  \includegraphics[width=0.9\linewidth]{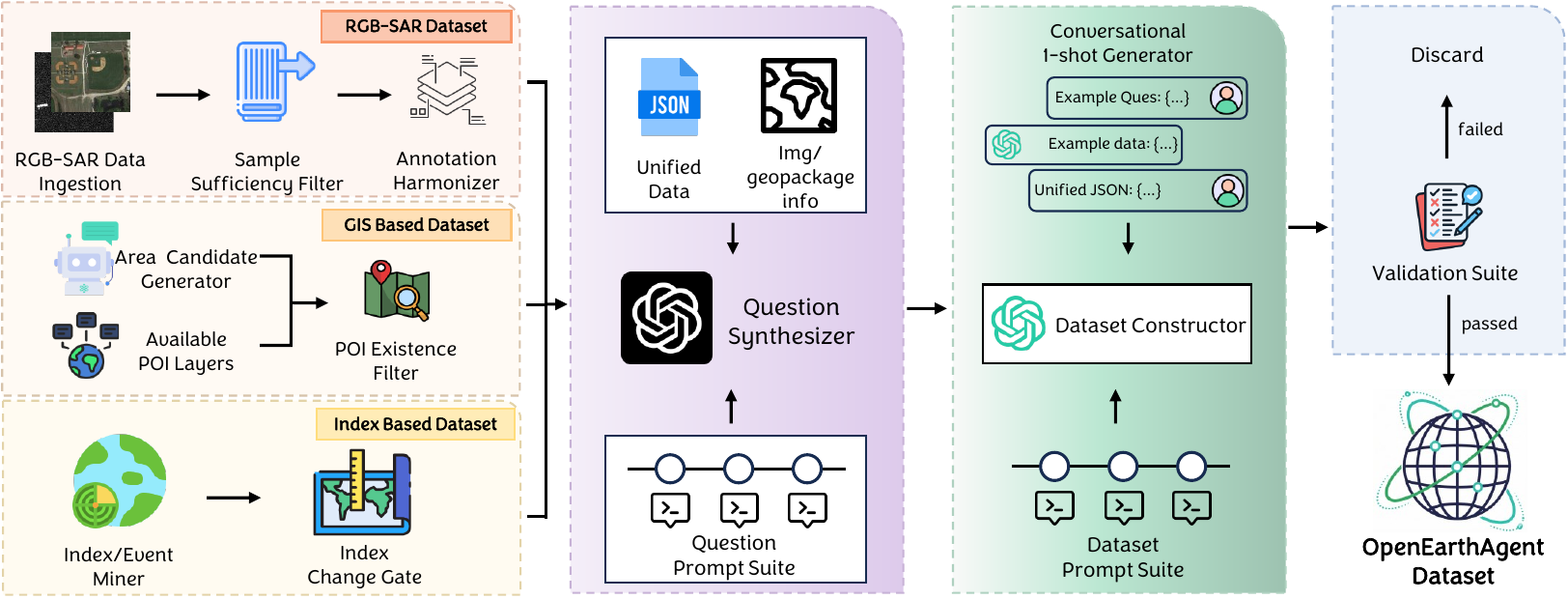}
  \caption{
  Unified data-curation pipeline for the OpenEarthAgent dataset. RGB/SAR imagery, GIS spatial layers, and index-driven datasets are processed through dedicated filtering (annotation checks, POI constraints, and spectral-change signals such as NDVI/NBR/NDBI). The resulting samples are merged into unified JSON records with imagery metadata and geopackage information. A question synthesizer and conversational generator produce tool-grounded queries and reasoning traces, which are automatically validated to ensure geographic and syntactic correctness before forming the final corpus for agentic training and evaluation.
  } 
  \vspace{-0.5em}
  \label{fig:data}
\end{figure*}
\begin{figure*}[t]
  \centering
  \includegraphics[width=\linewidth]{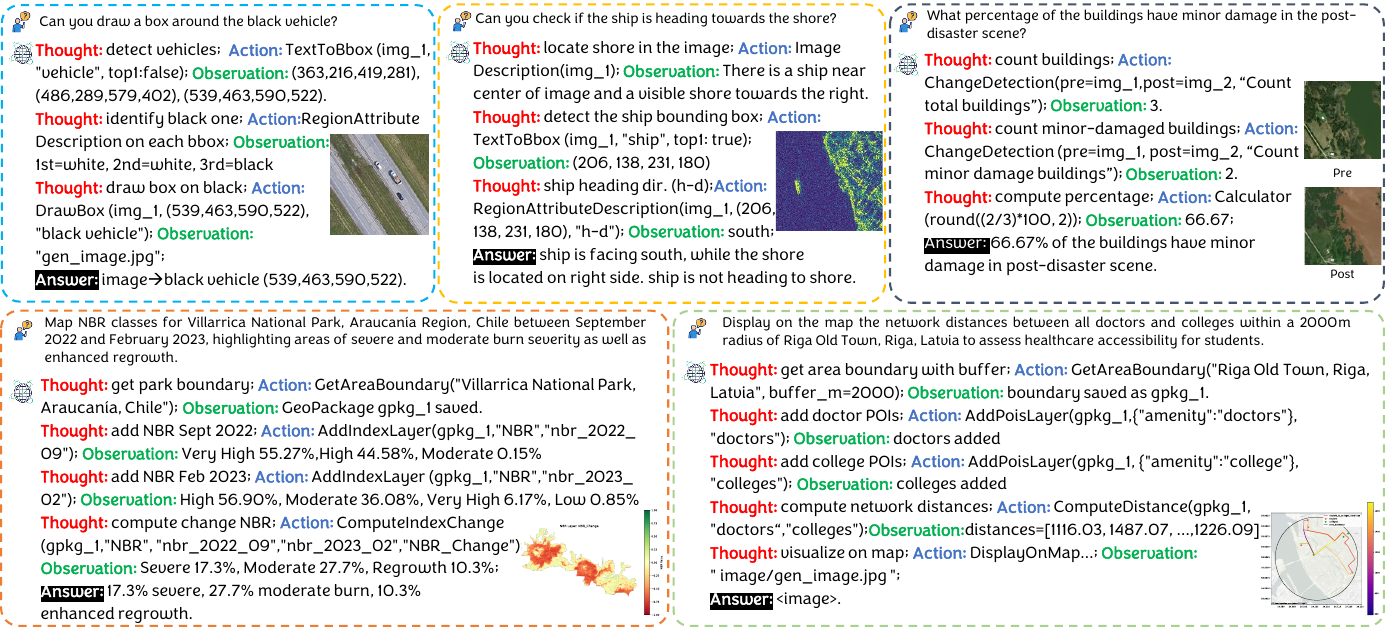}
  \caption{
  Representative query-reasoning trajectories from dataset, showing interleaving of natural-language thoughts, tool executions, and observations across tasks such as object localization, direction estimation, spectral mapping, change detection, and GIS-based analysis. Examples are \textit{simplified excerpts} of longer multi-step reasoning traces.
  } 
  \label{fig:examples}
\end{figure*}
\begin{figure}[!ht]
  \centering
  \begin{subfigure}[c]{0.48\linewidth}
    \centering
    \includegraphics[width=\linewidth]{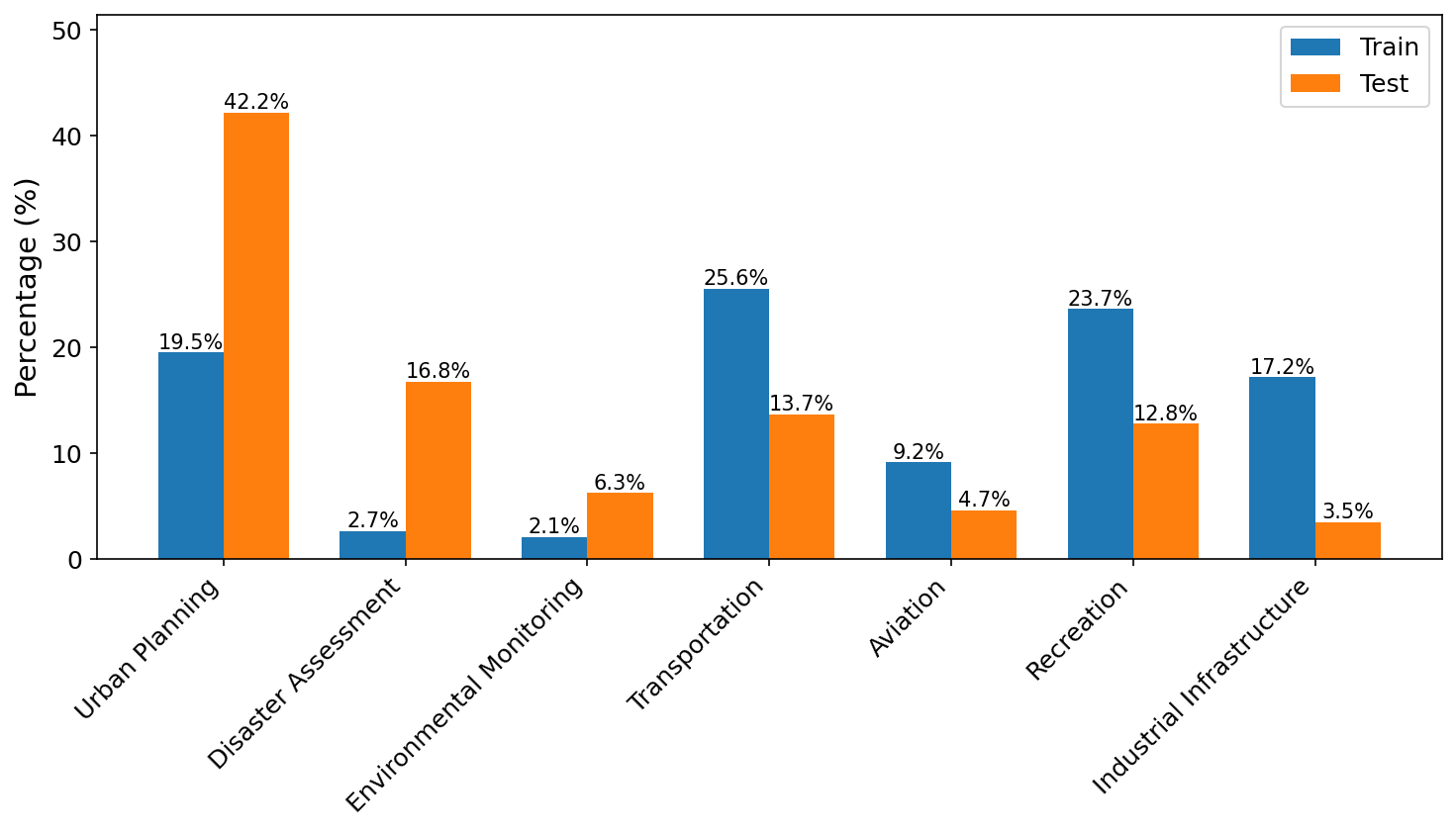}
    \caption{Data distribution across categories.}
    \label{fig:data_dist}
  \end{subfigure}\hfill
  \begin{subfigure}[c]{0.48\linewidth}
    \centering
    \includegraphics[width=\linewidth]{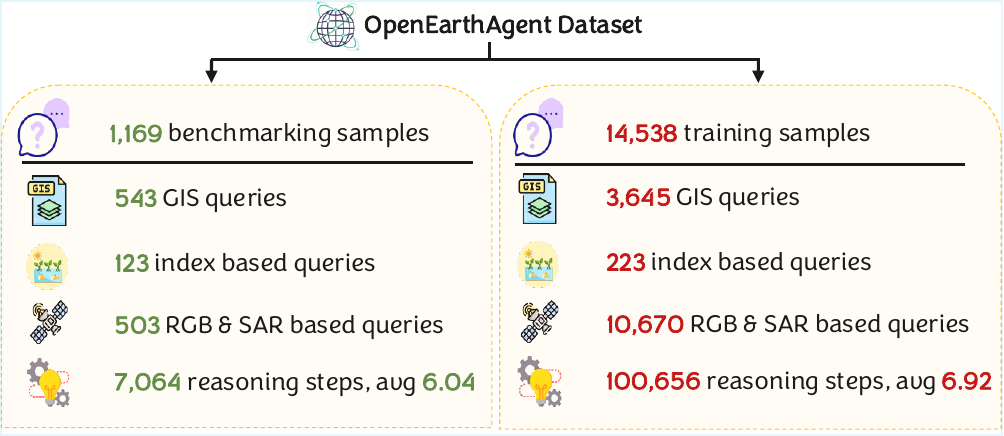}
    \vspace{0.05em}
    \caption{Key statistics for the curated corpus used for training and evaluation of our proposed agentic pipeline.}
    \label{fig:data_stats}
  \end{subfigure}
  \caption{Dataset overview: category distribution (left) and corpus statistics (right).}
  \label{fig:stats}
\end{figure}

\noindent\textbf{Automated Curation Pipeline and Quality Control:}
Dataset construction begins by aggregating candidate regions, annotations, and metadata from multiple geospatial sources.
For RGB-SAR branches, a \emph{Sample Sufficiency Filter} module removes datapoints that do not meet the required annotation counts, after which an \emph{Annotation Harmonizer} module aligns labels across source datasets to produce a unified annotation format.
Index-based samples are mined by detecting temporal variations in spectral indicators such as vegetation loss, burn severity, or flooding.
All sources are converted into a unified JSON schema encapsulating imagery paths, GeoPackage metadata, spatial attributes, and tool arguments.
Before the final generation, candidate GIS queries are programmatically executed (e.g., POI counting, distance computation, area estimation) to verify argument correctness and prevent formatting inconsistencies.
Potential inconsistencies such as mismatches between query instructions and tool parameters or invalid execution outputs are automatically detected, followed by controlled regeneration and targeted manual verification when necessary.

Natural-language queries and reasoning trajectories are synthesized through a structured LLM-driven module, with task-specific prompt templates and employ one-shot exemplars tailored to each reasoning category.
The data generator produces complete entries containing queries, multimodal inputs, and serialized reasoning traces.
Each instance is validated through corresponding tool execution to verify argument correctness, geometric validity, and spatial consistency.
All test-set samples undergo manual inspection to ensure realistic geospatial measurements (e.g., addressing GSD inconsistencies in area estimates).
Validated samples form the final corpus.
A summary of the pipeline is shown in Fig.~\ref{fig:data}.

\noindent\textbf{Task Structure and Query Design:}
Each instance represents a query-driven reasoning process that transforms static observation into an iterative perception-action loop (see Fig.~\ref{fig:examples}).
Queries are expressed in natural language and paired with structured reasoning traces that interleave \textit{thoughts}, \textit{actions}, and \textit{observations}, ultimately converging on a final \textit{answer}.
These traces span diverse modalities and task types, including:
GIS operations (distance, buffer, area, zonal statistics), object-level reasoning (counting, comparison, orientation, proximity), spectral index computation (NDVI, NBR, NDBI), and temporal change detection.
Reasoning depth and tool sequence length vary across samples, but each trajectory provides a complete chain of intermediate steps, enabling transparent multi-step geospatial problem solving rather than one-shot prediction.

\noindent\textbf{Dataset Statistics:}
Fig.~\ref{fig:stats} summarizes the scale, distribution, and structure of the OpenEarthAgent dataset.
The training split contains 14,538 instances, and the benchmark split contains 1,169 evaluation samples, each with multimodal imagery, a natural-language query, and a structured reasoning trace with interleaved tool calls and intermediate observations.
The training set contains 100,656 reasoning steps (average 6.92 per query), and the benchmark split contains 7,064 reasoning steps (average 6.04 per query).
Reasoning steps correspond to explicit thought-action-observation transitions, while tool calls capture the diversity of operational pathways required for solving curated agentic tasks.
The breadth of modalities, thematic coverage, and depth of reasoning traces enable rigorous evaluation of grounded, interpretable, and tool-augmented geospatial reasoning.

\begin{figure}[h]
  \centering
  \includegraphics[width=0.95\linewidth]{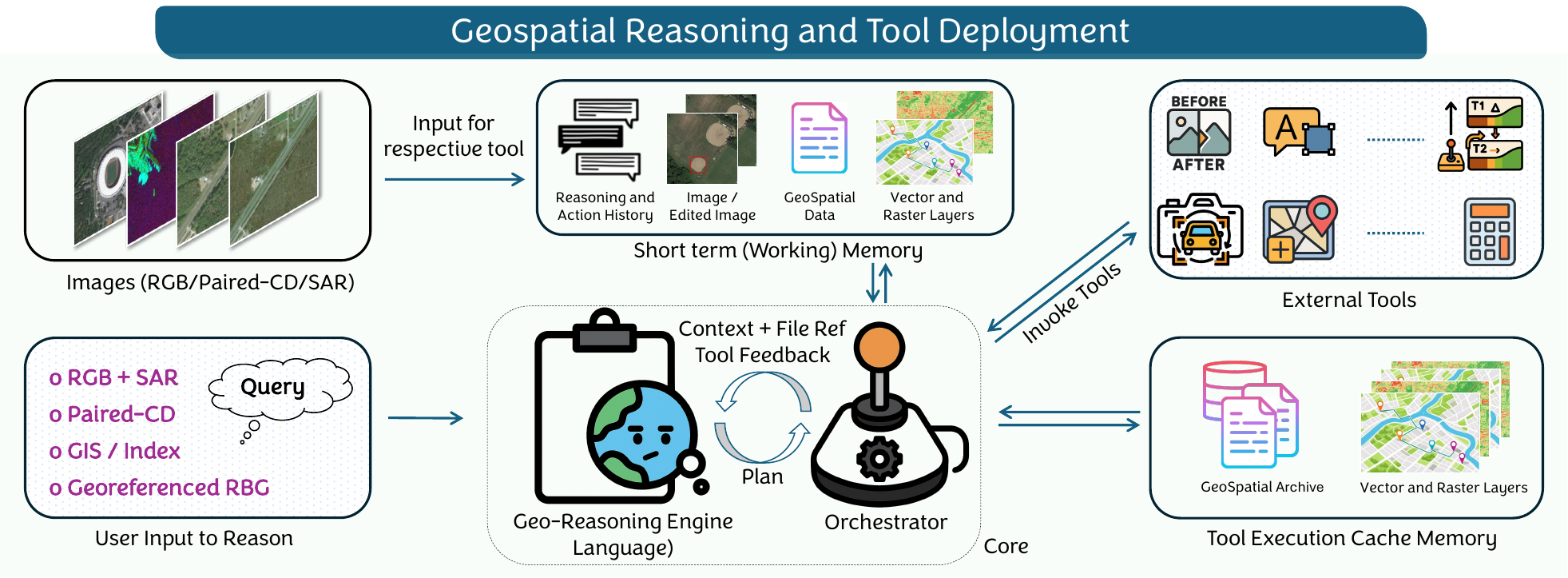}
  \caption{Overview of the proposed \textit{OpenEarthAgent} framework.
    The figure depicts tool deployment, where user queries over RGB, SAR, paired change-detection (CD), or GIS/indexed imagery are processed by the geospatial reasoning engine and tool orchestrator, which invoke appropriate tools and integrate feedback.} 
\label{fig:tool_fig}
\end{figure}

\subsection{Methodology}
\noindent\textbf{Overview:}
We present \textit{OpenEarthAgent}, a unified agentic learning framework for multimodal reasoning over earth observation data.
The framework integrates tool-driven interaction and language-grounded reasoning into a single end-to-end formulation.
Unlike conventional, single-step multimodal models, OpenEarthAgent decomposes geospatial reasoning into an organized sequence of perception and action steps, where the agent learns to \emph{perceive}, \emph{reason}, and \emph{act} through callable tools operating on visual, textual, and spatial modalities.
Given a natural-language instruction $\mathbf{u}$ and corresponding multimodal inputs $\mathbf{v}$ (e.g., RGB imagery, paired change-detection inputs, SAR, GIS layers, or georeferenced raster data), the agent generates multi-step reasoning trajectories that interleave internal thoughts with explicit tool invocations and their returned outputs (see Fig.~\ref{fig:tool_fig}). 
At each step, the agent maintains a short-term working memory that contains: instructions, observations, prior tool calls, spatial metadata, and contextual feedback from previous executions, which enable iterative reasoning and consistent spatial grounding.
Training follows a supervised fine-tuning procedure on validated reasoning traces, improving syntactic validity, spatial consistency, and multi-step reasoning coherence across diverse geospatial contexts.

\noindent\textbf{Unified Tool Registry:}
To standardize the diverse operators used in EO reasoning, we construct a unified tool registry that abstracts heterogeneous visual, spectral, and GIS operations through a consistent callable schema:
\begin{equation}
\mathcal{M}_j = (\mathbf{x}_{\text{in}}, \mathbf{y}_{\text{out}}, \psi_j),
\end{equation}
where $\mathbf{x}_{\text{in}}$ denotes structured input arguments, $\mathbf{y}_{\text{out}}$ denotes structured outputs, $\psi_j : \mathbf{x}_{\text{in}} \mapsto \mathbf{y}_{\text{out}}$ is the executable function implementing the operation.
All tools follow standardized JSON-based contracts to ensure consistent serialization and validation across training and inference.
A central orchestrator parses model-generated tool calls, validates arguments, executes $\psi_j$, and appends returned outputs to the working memory.
During execution, a predicted tool invocation $s_t$ with arguments $\hat{\mathbf{x}}_t$ is evaluated as $\mathbf{y}_t = \psi_{s_t}(\hat{\mathbf{x}}_t)$,
and the returned output $\mathbf{y}_t$ becomes part of the updated reasoning context.

To ensure efficiency and consistency, intermediate outputs (e.g., derived vector layers, raster subsets, index maps, computed geometries) are stored in a tool execution cache memory.
This cache includes geospatial archives and vector/raster layers that can be reused by subsequent tool calls within the same trajectory.
Caching prevents redundant computation and ensures deterministic replay of reasoning steps.
New tools can be integrated by registering their schema $\mathcal{M}_j$ without retraining the backbone model, ensuring extensibility while maintaining interface consistency.
We outline the tool categories:

\noindent\textbf{a) Perceptual Tools.}
ground language into image-space entities.
\texttt{Text\allowbreak To\allowbreak Bbox}, \texttt{Object\allowbreak Detection}, and \texttt{Region\allowbreak Attribute\allowbreak Description} convert free-form queries into spatial primitives such as bounding boxes and semantic tags.
\texttt{Count\allowbreak Given\allowbreak Object} and \texttt{Segment\allowbreak Object\allowbreak Pixels} support object counting and pixel-level area estimation.
\texttt{Change\allowbreak Detection} analyzes multi-date imagery to detect structural changes such as damage or flooding.
\noindent\textbf{b) GIS Computation Tools.} enable geographic reasoning via geometric and coordinate-aware computation.
\texttt{GetArea\allowbreak Boundary}, \texttt{AddPoisLayer}, and \texttt{ComputeDistance} manipulate vector geometries within standard coordinate reference systems.
Outputs include numerical measures (e.g., distances, buffers, zonal statistics) or derived layers visualized via \texttt{DisplayOnMap}.
\noindent\textbf{c) Spectral Tools.} perform band arithmetic and spectral analysis on raster inputs.
\texttt{AddIndexLayer}, \texttt{ComputeIndexChange}, and \texttt{ShowIndex\allowbreak Layer} compute and visualize indices (NDVI/NBR/NDBI), enabling quantitative reasoning about environmental dynamics.
\noindent\textbf{d) Georeferenced Raster Tools.} \texttt{Get\allowbreak Bbox\allowbreak From\allowbreak Geotiff} and \texttt{Display\allowbreak On\allowbreak Geo\allowbreak tiff} operate directly on georeferenced rasters supporting coordinate extraction, bounding box derivation, and visualization aligned with spatial metadata, ensuring spatial consistency.
\noindent\textbf{e) Utility Tools.}
\texttt{Calculator}, \texttt{Solver}, and \texttt{Plot} support arithmetic and symbolic reasoning.
Visualization tools \texttt{DrawBox} and \texttt{AddText} display results.
\texttt{OCR} and \texttt{GoogleSearch} extend multimodal understanding through text extraction and contextual retrieval.
\texttt{Terminate} signals completion of a reasoning trajectory.

\noindent\textbf{Reasoning Trajectories and Training Objective:}
Each training sample encodes a reasoning trajectory that connects an instruction $\mathbf{u}$, observations $\mathbf{v}$, and a sequence of tool interactions.
Formally, the trajectory is represented as:
\begin{align}
\Gamma_i &= \{(s_t, r_t)\}_{t=1}^{T_i}, \quad s_t \in \mathcal{S}, \nonumber\\
r_t &\sim \mathcal{E}\!\big(s_t \mid \mathbf{u}_i, \mathbf{v}_i, \{(s_k, r_k)\}_{k<t}\big),
\end{align}
where $\mathcal{S}$ denotes the set of registered tools, $\mathcal{E}$ the execution environment (tool controller + cache), and $(s_t, r_t)$ is the $t$-th action-result pair.
Each $s_t$ is a predicted tool invocation, while $r_t$ is the observation returned by the environment.

The autoregressive model conditions on the full working memory state when predicting the next action.
Training is performed via maximum likelihood estimation over verified trajectories, optimizing only the tool-action policy:

\begin{equation}
\mathcal{L}_{\text{train}}
= -\frac{1}{N}\sum_{i=1}^{N}\sum_{t=1}^{T_i}
\log P_\eta(s_t \mid \mathbf{u}_i, \mathbf{v}_i, s_{<t}, r_{<t}),
\end{equation}

where $\eta$ parameterizes the model and $N$ is the number of training samples.
Tool observations $r_t$ are added to the working memory as context but are masked from loss computation and not treated as policy outputs.
This formulation enforces syntactic validity of tool calls, sequential and spatial coherence, and argument correctness, while preserving the environment-policy separation.

Before inclusion in training, each trajectory undergoes deterministic replay through the tool controller.
Re-execution verifies argument formatting, coordinate integrity, geometric validity, and full-chain executability.
Through supervised fine-tuning on validated multi-step trajectories, OpenEarthAgent learns to generate syntactically correct, spatially grounded, and interpretable reasoning workflows over heterogeneous EO data.

\section{Experiments}
In this section, we outline implementation details, describe the two evaluation modes, define metrics used to quantify reasoning fidelity, tool effectiveness, and geospatial accuracy, and discuss key experimental results and insights.

\noindent\textbf{Implementation Details: }
OpenEarthAgent is finetuned using Qwen3-4B-Instruct-2507 as the base model. Training was performed on LLM controller on four NVIDIA A100 GPUs (40 GB). The base model is loaded via Unsloth's FastLanguageModel, enabling efficient multi-GPU distributed training. The model is trained for one epoch with a learning rate of $2\times10^{-5}$, cosine learning rate scheduling, 0.05 warmup ratio, and a batch size of 16. The maximum sequence length is set to 4096 tokens to support long-context conversational tool interactions. 
LLM is finetuned on conventional data, where the model alternates between text generation and tool invocation, providing tool outputs and feedback in JSON format, embedded in its user and assistant turns. Training applies response-only masking, computing loss exclusively on assistant-generated tokens, enabling accurate tool calls and structured responses without optimizing on prompt text or external tool outputs. This enables the model to establish a good grounding for tool invocation. 

\subsection{Baseline Evaluation}

To evaluate OpenEarthAgent, we benchmark its performance against a diverse set of large language models (LLMs), encompassing both proprietary and open-source variants. The comparison includes advanced frontier models (GPT-4o and its reasoning-optimized variant o4-mini), and mid- to small-scale open-weight models, including Llama3.1 and Internlm3.
All models serve as interchangeable backbones within the unified OpenEarthAgent framework. We assess performance using both step-by-step and end-to-end protocols, allowing analysis of intermediate reasoning trajectories and final task completion accuracy.

\noindent\textbf{Step-by-Step Evaluation:} This mode measures procedural reasoning and tool-invocation behavior without executing tools. The model is given $n$ reasoning steps and must generate a valid action at each one based on the full interaction history. The first step is exempt from validation to allow the model to formulate an initial high-level plan before committing to concrete tool interactions. This setup isolates reasoning quality and enables focused analysis of spatial semantics, plan coherence, and geospatial context understanding. 

\noindent\textbf{End-to-End Evaluation:} This mode tests full autonomous execution with live tool use. The model issues tool calls, forms arguments, and reasons step-by-step based on prior outputs. This captures operational robustness, error handling, and the effectiveness of linking perception to action in geospatial workflows.

\setlength{\tabcolsep}{5pt}
\begin{table}[b!]
\scriptsize
\centering
\caption{Step-by-step evaluation under tool-agnostic rollouts. We report accuracies: Inst.; syntactic + logical validity of actions, Tool.; correct tool selection, ArgN.; presence of required parameters, ArgV.; parameter correctness, and Summ.; correctness of the final consolidated answer. While frontier LLMs lead on summary accuracy, OpenEarthAgent attains state-of-the-art Inst./Tool./ArgN./ArgV., narrowing the gap despite a substantially smaller parameter budget.}
\label{tab:model_performance}
\begin{tabular}{l@{\hspace{0pt}}c@{\hspace{6pt}}c@{\hspace{6pt}}c@{\hspace{6pt}}c@{\hspace{6pt}}c@{\hspace{6pt}}c}
\hline
\textbf{Model} &\textbf{Param.}& \textbf{Inst.} & \textbf{Tool.} & \textbf{ArgN.} & \textbf{ArgV.} & \textbf{Summ.} \\
\hline
\rowcolor{OpticalBlue!60}
\multicolumn{7}{c}{\textsc{FrontierLLMs}} \\
gpt-5 & & 97.54 & 82.69 & 73.28 & 37.28 & 87.02 \\
gpt-4o & & 99.18 & 93.88 & 85.48 & 45.80 & 86.76 \\
o4-mini & & 83.68 & 68.33 & 64.17 & 37.95 & \textbf{89.48} \\
\midrule
\rowcolor{OpticalBlue!60}
\multicolumn{7}{c}{\textsc{Open-SourceLLMs}} \\
Qwen2.5-Instruct &7B& 94.08 & 85.51 & 78.46 & 38.00 & 80.08 \\
Llama-3.1-Instruct  &8B& 47.07 & 39.30 & 34.11 & 17.49 & 79.08 \\
Internlm3-Instruct &8B& 44.54 & 38.46 & 27.82 & 13.25 & 29.84 \\
Mistralv0.3-Instruct  &7B& 64.14 & 35.12 & 26.11 & 12.45 & 24.04 \\
Qwen2.5-Instruct  &3B& 85.07 & 72.13 & 64.87 & 24.12 & 68.75 \\
\midrule
\rowcolor{OpticalBlue!60}
\multicolumn{7}{c}{\textsc{BaselineLLM}} \\
Qwen3-Instruct-2507  &4B& 97.34 & 86.94 & 84.12 & 33.55 & 83.28 \\
\midrule
\rowcolor{OpticalBlue!60}
\multicolumn{7}{c}{\textsc{OpenEarthAgent}} \\
OpenEarthAgent &4B& \textbf{99.51} & \textbf{97.18} & \textbf{96.08} & \textbf{62.10} & 83.64 \\
\hline
\end{tabular}
\end{table}
\setlength{\tabcolsep}{6pt}

\begin{table}[t]
\small
\centering
\caption{End-to-end evaluation: (i) F$_1$ scores for tool selection across: Perception (Per.), Operation (Op.), Logic (Logic.), and GIS (GIS.); (ii) tool-order fidelity: \emph{AnyOrder} (multiset match, order-agnostic), \emph{SameOrder} (exact sequence match), and \emph{Unique} (set match, no multiplicity); and (iii) task-level Accuracy, \emph{Ans.} for non-generative and \emph{Gen.} for image-generation. GPT-4o attains high GIS F$_1$, while OpenEarthAgent achieves strong, balanced performance with superior trajectory fidelity.
}
\label{tab:e2e_metrics}
\resizebox{\textwidth}{!}{%
\begin{tabular}{lcccccccccc}
\hline
\multirow{2}{*}{\textbf{Model}}& \multirow{2}{*}{\textbf{Param.}}&
\multicolumn{4}{c}{\textbf{F$_1$ scores}} &
\multicolumn{3}{c}{\textbf{Tool Order}} &
\multicolumn{2}{c}{\textbf{Accuracy}} \\
\cmidrule(lr){3-6}\cmidrule(lr){7-9}\cmidrule(lr){10-11}
 &&
\textbf{Per.} &
\textbf{Op.} &
\textbf{Logic.} &
\textbf{GIS.} &
\textbf{AnyOr.} &
\textbf{SameO.} &
\textbf{Uni.} &
\textbf{Ans.} &
\textbf{Gen.} \\
\hline
\rowcolor{OpticalBlue!60}
\multicolumn{11}{c}{\textsc{FrontierLLMs}} \\
gpt-5 & & 16.88 & 47.00 & 6.26 & 91.50 & 46.96 & 46.79 & 47.81 & 43.88 & 46.21 \\
gpt-4o & & 44.47 & \textbf{66.91} & 35.95 & 95.80 & 50.81 & 50.38& 55.52 & 39.22 & \textbf{77.93} \\
o4-mini & & 39.48 & 64.93 & 12.38 & 86.51 & 40.12 & 39.95 & 41.49 & 35.18 & 55.17 \\
\midrule
\rowcolor{OpticalBlue!60}
\multicolumn{11}{c}{\textsc{Open-SourceLLMs}} \\
Qwen2.5-Instruct &7B &20.38 & 37.33 & 26.68 & 75.46 & 31.57 & 30.02 & 36.61 & 15.85 & 41.38 \\
Llama-3.1-Instruct & 8B& 22.54 & 16.45 & 32.03 & 64.64 & 39.01 & 37.72 & 44.91 & 12.70 & 55.17  \\
Internlm3-instruct & 8B&5.50 & 14.21  & 0.51 & 31.47 & 2.22 & 1.88 & 3.16 & 0.24 & 2.76 \\
Mistral-Instruct-v0.3 & 7B & 6.74 & 10.24 & 15.82 & 17.12 & 2.65  & 2.57 & 2.91 & 0.39 & 5.52 \\
Qwen2.5-Instruct &3B &15.63 & 16.54 & 11.16 & 35.64 & 9.24 & 8.72 & 12.40 & 1.83 & 24.14 \\
\midrule
\rowcolor{OpticalBlue!60}
\multicolumn{11}{c}{\textsc{BaselineLLM}} \\
Qwen3-Instruct-2507 & 4B&13.86 & 20.95 & 18.35 & 71.82 & 16.00 & 14.71 & 21.47 & 13.72 & 15.86 \\
\midrule
\rowcolor{OpticalBlue!60}
\multicolumn{11}{c}{\textsc{OpenEarthAgent}} \\
OpenEarthAgent &4B &\textbf{58.30} & 56.76 & \textbf{51.18} & \textbf{98.52} & \textbf{67.75} & \textbf{67.24} & \textbf{72.71} & \textbf{45.26} & 75.86 \\
\hline
\end{tabular}}
\end{table}

\noindent\textbf{Evaluation Metrics:} 
We assess the model’s procedural reasoning with five metrics. Instance Accuracy (Inst.) measures the proportion of tool called without logical or syntactic errors; Tool Accuracy (Tool) evaluates correctness of tool selection; Argument Name Accuracy (ArgN) checks the presence of all required arguments in the generated action; Argument Value Accuracy (ArgV) verifies argument correctness; and Summarization Accuracy (SummAcc) quantifies how well the model consolidates information from prior tool calls into a coherent final response.
We compute F$_1$ for tool selection across four functional categories: Perception (Per.), Operation (Op.), Logic (Logic.), and GIS (GIS.) to capture alignment between predicted and reference tool sets. Trajectory quality is further assessed via tool-order accuracy under increasing constraints: \textit{Unique} checks tool-to-tool correspondence without penalizing repeated calls, \textit{AnyOrder} requires all tools (including duplicates) regardless of order, and \textit{SameOrder} enforces both correct sequence and frequency of tool invocations. Overall task Accuracy compares final model completions with reference outputs. For image-generation tasks, success is determined by the correctness of generation tool calls and their execution success rate; all non-generation queries are evaluated by an LLM judge (gpt-4o-mini) using a standardized elaboration evaluation prompt.

\subsection{Results}
In step-by-step evaluation (Table \ref{tab:model_performance}), frontier models lead: o4-mini score highest on Summ.\ (89.48) while gpt-4o shows strong Inst. Acc. Among open-source models, Qwen2.5-I (7B) performs competitively (Inst. 94.08, Tool. 85.51), whereas other 7B–8B models degrade, particularly on ArgV. prediction. OpenEarthAgent (4B) substantially narrows this gap despite its smaller size, achieving state-of-the-art Inst./ Tool./ ArgN./ and ArgV.(99.51/97.18/96.08/62.10). In the end-to-end evaluation (Table \ref{tab:e2e_metrics}), GPT-5 avoids external logic tools and relies on internal numerical and analytical computations. GPT-4o leads in Op.\ F$_1$ (66.91) and Gen. Acc. (77.93) but remains modest in Per./Logic, whereas OpenEarthAgent delivers balanced gains across Per./Op./Logic./GIS and dominates in tool-order accuracy (AnyOr./SameO. /Uni. $\approx$ 67-72\%), while substantially improving Answer Acc. (45.26) indicating robust trajectory planning rather than isolated tool correctness. Overall, these results suggest that targeted training with tool schemas and trajectories can outperform larger general-purpose models for geospatial tool use, especially under strict order/frequency constraints.

\noindent\textbf{Cross-Benchmark Evaluation} We extend our evaluation to Earth-Agent ~\cite{feng2025earth} benchmark(248 tasks, 13,729 images spanning the spectrum, products, and RGB modalities), scoring with fine-grained step-level metrics. For compatibility, benchmark tools are reformatted to the OpenEarthAgent interface, where curated JSON exemplars standardize tool invocation. As Table~\ref{tab:cross_bm_res} reports, OpenEarthAgent (4B) outperforms similarly sized open-source baselines on every tool-related metric, and rivals GPT-4o, showing that structured agent design and training bring small models to within reach of frontier-level step-wise geospatial reasoning and tool execution. These gains transfer to ThinkGeo, evident from Table ~\ref{tab:gpt_agent_comparison} (Left), OpenEarthAgent attains the highest Ans (16.35) and Ans\_I (29.13), surpassing both the reported ThinkGeo results and the Qwen3-4B baseline. Such consistency across benchmarks confirms that OpenEarthAgent couples reasoning and tool use into a robust, transferable capability.

\setlength{\tabcolsep}{5pt}
\begin{table}[t!]
\scriptsize
\centering
\caption{Step-by-step evaluation on Earth-Agent Benchmark. }
\label{tab:cross_bm_res}
\begin{tabular}{l@{\hspace{0pt}}c@{\hspace{6pt}}c@{\hspace{6pt}}c@{\hspace{6pt}}c@{\hspace{6pt}}c@{\hspace{6pt}}c}
\hline
\textbf{Model} &\textbf{Param.}& \textbf{Inst.} & \textbf{Tool.} & \textbf{ArgN.} & \textbf{ArgV.} & \textbf{Summ.} \\
\hline
\rowcolor{OpticalBlue!60}
\multicolumn{7}{c}{\textsc{FrontierLLMs}} \\
gpt-4o & & \textbf{97.41} & \textbf{71.65} & \textbf{70.01} & \textbf{48.11} & \textbf{80.26} \\
\midrule
\rowcolor{OpticalBlue!60}
\multicolumn{7}{c}{\textsc{Open-SourceLLMs}} \\
Qwen2.5-Instruct &7B& 85.98 & 57.76 & 56.06 & 34.66 & \underline{79.60} \\
Llama-3.1-Instruct  &8B& 83.21 & 55.11 & 54.29 & 35.98 & 73.05 \\
\midrule
\rowcolor{OpticalBlue!60}
\multicolumn{7}{c}{\textsc{BaselineLLM}} \\
Qwen3-Instruct-2507  &4B& 88.32 & 60.61 & 59.53 & 38.89 & 76.79 \\
\midrule
\rowcolor{OpticalBlue!60}
\multicolumn{7}{c}{\textsc{OpenEarthAgent}} \\
OpenEarthAgent &4B& \underline{96.91} & \underline{70.33} & \underline{69.13} & \underline{44.44} & 79.33 \\
\hline
\end{tabular}
\end{table}
\setlength{\tabcolsep}{6pt}

\begin{table}[t]
\centering
\caption{(Left) ThinkGeo benchmark comparison on answer-level metrics. OpenEarthAgent achieves the scores, outperforming ThinkGeo and Qwen3-4B. (Right) Comparison with a generic GPT agent across the spectrum, product, and image tasks. OpenEarthAgent achieves the best overall performance with the lowest latency.}
\label{tab:gpt_agent_comparison}
\scriptsize
\makebox[\columnwidth][c]{%
\begin{tabular}{@{}p{2.7cm}@{} p{.6cm} p{.9cm}@{}}
\toprule
\textbf{Model} & \textbf{Ans} & \textbf{Ans\_I} \\
\midrule
ThinkGeo (reported) & 9.78 & 20.40 \\
Qwen3-4B (BL) & 12.53 & 25.92 \\
\rowcolor{OpticalBlue!60}
OpenEarthAgent & \textbf{16.35} & \textbf{29.13} \\
\bottomrule
\end{tabular}
\hspace{1.3em}
\begin{tabular}{@{}p{2.5cm}@{} p{.5cm} p{.5cm} p{.5cm} p{.5cm} p{.9cm}@{}}
\toprule
\textbf{Model} & \textbf{Spec.} & \textbf{Prod.} & \textbf{Img.} & \textbf{Total} & \textbf{Lat(s)} \\
\midrule
GPT-Agent & 14.67 & 22.25 & 28.33 & 21.75 & 933.54 \\
Qwen3-4B (BL) & 31.07 & 13.98 & 23.66 & 22.90 & 42.02 \\
\rowcolor{OpticalBlue!60}
OpenEarthAgent & \textbf{90.46} & \textbf{60.67} & \textbf{29.00} & \textbf{60.04} & \textbf{21.67} \\
\bottomrule
\end{tabular}%
}
\end{table}
\noindent\textbf{Comparison with General Agent} To assess OpenEarthAgent against a general purpose agent, we evaluate a balanced 60-query subset(20 spectrum, 20 product, and 20 image-analysis tasks) spanning the benchmark's full range of reasoning and tool-use demands. Table~\ref{tab:gpt_agent_comparison} (Right) shows OpenEarthAgent leading in every category, with the widest margins on spectrum tasks, where it scores 90.46, more than 4 times GPT-Agent's 14.67. Overall, OpenEarthAgent achieves 60.04 while \texttt{GPT-Agent} reaches only 21.75, and the \texttt{Qwen3-4B} baseline reaches 22.90. OpenEarthAgent is also far more efficient, averaging just 21.67 seconds per query versus 42.02 seconds for Qwen3-4B (\~2x slower) and 933.54 seconds for GPT-Agent (\~44x slower).  OpenEarthAgent's task-specific tool selection yields higher accuracy and lower latency than generic problem-solving.

\begin{figure*}[t]
    \centering

    \begin{minipage}[t]{0.48\linewidth}
        \centering
        \includegraphics[width=\linewidth]{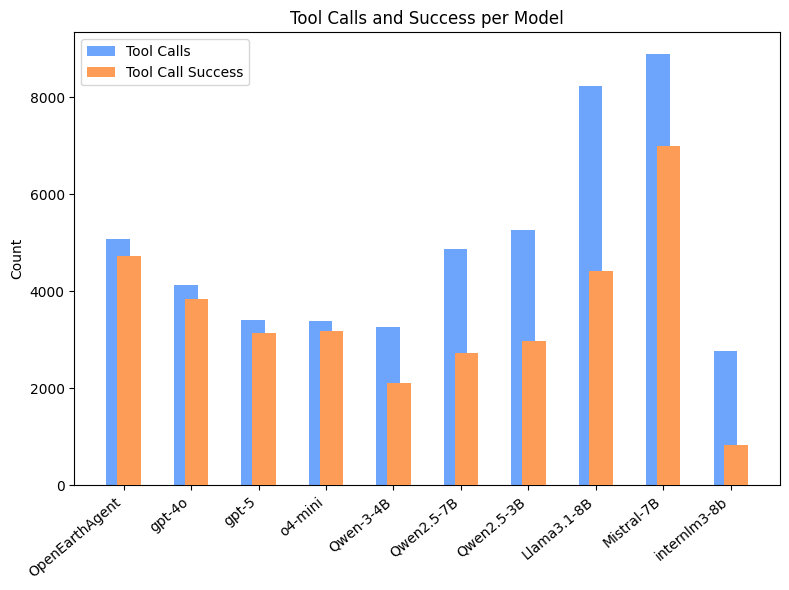}
        \caption{Tool-call performance: Open-source models show lower success rates, while GPT family and OpenEarthAgent achieve substantially higher success rates.}
        \label{fig:fig1}
    \end{minipage}
    \hfill
    \begin{minipage}[t]{0.48\linewidth}
        \centering
        \includegraphics[width=\linewidth]{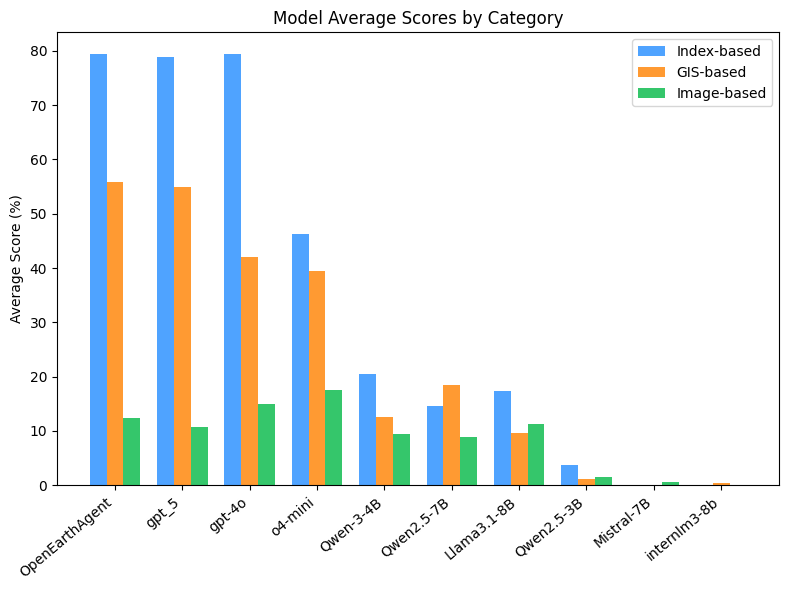}
        \caption{Category-wise performance: OpenEarthAgent and gpt-5 lead in index/ GIS score, while open source performs comparatively on image-based queries.}
        \label{fig:fig2}
    \end{minipage}

\end{figure*}

\noindent\textbf{Tool Call \& Success Patterns:} 
Proprietary models (gpt-5, gpt-4o and o4-mini) show strong tool-control, issuing thousands of calls with low error rates ($\sim$7\%), indicating stable formatting. OpenEarthAgent executes a higher volume of tool calls while achieving a superior overall success rate. In contrast, open-source models such as Qwen2.5-7B, Qwen2.5-3B, and Llama-3.1-8B exhibit high failure rates (43-46\%), reflecting weak adherence to the tool schema. Mistral-7B demonstrates a lower error rate but fails to invoke tools at the required adequacy to meet task demands. Despite extensive tool usage, Llama-3.1-8B and Mistral-7B fail to converge to a final answer within the allowed step, resulting in 94\% and 98\% incomplete task executions, respectively. Overall, performance as shown in  Fig.~\ref{fig:fig1} varies by call volume and precision: robust models maintain low error rates, while others struggle with consistent, well-formed tool invocations.

\noindent\textbf{Category Scores:}
As seen in Fig.~\ref{fig:fig2}, OpenEarthAgent (79.43\%), gpt-4o (79.39 \%), and gpt-5 (78.94\%) achieve the strongest index-based scores, substantially outperforming all baselines. GIS-based tasks reveal a sharper degradation, with OpenEarthAgent (55.77 \%) followed by gpt-5 (54.94 \%), gpt-4o (41.95 \%), and o4-mini (39.55\%). Open-source models; qwen2.5-7b, qwen3-4b, and llama3.1-8b perform comparably well on image-based queries, while smaller models approach zero on index tasks and score only marginally on GIS and image-based tasks. Overall, the results highlight a significant performance gap and show that index- and GIS-based reasoning remains difficult for smaller open models. 
\noindent Additional analyses and ablations are provided in the supplementary material.

\section{Conclusion}
We introduced \textit{OpenEarthAgent}, a unified framework for tool-augmented geospatial reasoning that connects natural language with multimodal earth observation data through structured analytical workflows.
Unlike conventional EO models that focus primarily on perception, OpenEarthAgent performs grounded and interpretable reasoning by orchestrating visual, spectral, GIS, and GeoTIFF-aware tools under a consistent executable schema.
The accompanying corpus comprises 14,538 training instances and 1,169 held-out test instances with validated multi-step reasoning trajectories, including explicit tool calls, intermediate observations, and outputs.
OpenEarthAgent addresses a key gap in EO research by moving beyond perception-only modeling toward structured and verifiable geospatial reasoning.
We believe this work provides a step toward agentic AI systems for environmental monitoring, disaster response, infrastructure analysis, and broader geospatial decision-making.

\bibliographystyle{splncs04}
\bibliography{main}

\clearpage
\setcounter{page}{1}

\begin{center}
    {\large \textbf{Supplementary}}\\[1em]
    {\large OpenEarthAgent: A Unified Framework for Tool-Augmented Geospatial Agents}
\end{center}

\renewcommand{\thefigure}{A\arabic{figure}}
\setcounter{figure}{0}  
\renewcommand{\thetable}{A\arabic{table}}
\setcounter{table}{0}
\newcounter{secnumber}
\setcounter{secnumber}{1} 
\renewcommand{\thesection}{S\arabic{secnumber}}

\definecolor{OpticalBlue}{HTML}{E9F2FF}
\definecolor{SARGreen}{HTML}{E9FBE9}
\definecolor{GISOrange}{HTML}{FFF3E0}
\definecolor{IndexPurple}{HTML}{F3E9FF}

\noindent Supplementary material includes ablation analysis (\ref{sec:ablation}), model selection criterion (\ref{sec:modelselection}), error analysis (\ref{sec:erroranalysis}), and context-specific performance assessment (\ref{sec:contanalysis}). It also provides quantitative results illustrating multiple evaluation cases (\ref{sec:qualitative}), a detailed overview of input-output structures with corresponding tool descriptions and details (\ref{sec:tooldescription}), an evaluation prompt illustration with a comparative study on LLM Judge (\ref{sec:eval_prompt_sec}), and a detailed description of the dataset-curation components used to construct the OpenEarthAgent corpus (\ref{sec:dataset_curation_components}).

\section{Ablation Analysis}
\stepcounter{secnumber}
\label{sec:ablation}

\subsection*{Effect of Data Composition}
To investigate how data composition influences model performance, we divide the training set into easy and hard subsets based on the length of reasoning trajectories, measured by the number of assistant planning and action steps in each conversation. First, samples with fewer than seven tool-calling turns are categorized as easy, while those with seven or more turns are considered hard. To avoid tool-distribution bias, we further rebalanced several underrepresented tools by evenly redistributing their samples across the two subsets. The easy subset contains 8,113 samples averaging 6.095 turns, while the hard subset contains 6,425 samples averaging 7.975 turns. We evaluated under three conditions: training on easy-only, hard-only, and combined data. 
\begin{table*}[b!]
\small
\centering
\caption{Impact of data composition: Easy-only training achieves the best Perception score, Hard-only training boosts operational reasoning, while the combined setting yields the most balanced performance overall, achieving the highest Logic and GIS scores, the strongest tool-order consistency, and the best answer accuracy.}
\label{tab:abl1}
\resizebox{\textwidth}{!}{%
\begin{tabular}{lccccccccccc}
\hline
\multirow{2}{*}{\textbf{Data}}&\multicolumn{2}{c}{\textbf{Abl. setting}}&
\multicolumn{4}{c}{\textbf{F$_1$ scores}} &
\multicolumn{3}{c}{\textbf{Tool Order}} &
\multicolumn{2}{c}{\textbf{Accuracy}} \\
\cmidrule(lr){2-3}\cmidrule(lr){4-7}\cmidrule(lr){8-10}\cmidrule(lr){11-12}
&\textbf{Easy}&\textbf{Hard}&
\textbf{Per.} &
\textbf{Op.} &
\textbf{Logic.} &
\textbf{GIS.} &
\textbf{AnyO.} &
\textbf{SameO.} &
\textbf{Unique} &
\textbf{Ans.} &
\textbf{Gen.} \\
\midrule
\rowcolor{OpticalBlue!60}
\multicolumn{12}{c}{\textsc{OpenEarthAgent}} \\
Set1&{\textcolor{green!70!black}{$\checkmark$}}&\textcolor{red}{$\times$}& \textbf{59.63} & 52.60 & 48.11 & 97.26 & 61.59 & 61.42 & 67.51 & 43.40 & 73.79 \\
Set2&\textcolor{red}{$\times$}&{\textcolor{green!70!black}{$\checkmark$}}& 54.45 & \textbf{56.83} & 35.16 & 94.35 & 65.61 & 64.93 & 70.40 & 35.79 & 71.72 \\
Set1+Set2  &{\textcolor{green!70!black}{$\checkmark$}} & {\textcolor{green!70!black}{$\checkmark$}}& 58.30 & 56.76 & \textbf{51.18} & \textbf{98.52} & \textbf{67.75} & \textbf{67.24} & \textbf{72.71} & \textbf{45.26} & \textbf{75.86}\\
\hline
\end{tabular}}

\end{table*} 

Table~\ref{tab:abl1} analyzes how reasoning trajectory complexity (easy vs. hard) affects tool-selection F$_1$, trajectory consistency, and end-to-end task accuracy. Training on shorter trajectories (avg. 6.095 turns) yields the highest Perception F$_1$ (59.63) as perception tools (e.g., OCR, ObjectDetection, SegmentObjectPixels, CountGivenObject) are typically associated with early-stage spatial grounding and visual interpretation. Shorter trajectories, therefore, provide more focused supervision on these perceptual and semantic alignment steps. Training on longer trajectories (avg. 7.975 turns) enhances Operation F$_1$ (56.83) and improves tool-order robustness (AnyO 65.61, Unique 70.40). Hard samples involve extended reasoning chains and richer tool compositions, encouraging the model to learn procedural dependencies and structured multi-step workflows.
However, performance in logical reasoning (35.16) remains below the combined setting low, suggesting that while challenging samples strengthen structured reasoning, they benefit from complementary, simpler cases that help stabilize reasoning chains and reinforce logical consistency across steps.

The combined setting achieves the most balanced and strongest overall performance, yielding the highest Logic F$_1$ (51.18) and GIS F$_1$ (98.52), the strongest tool-order consistency (AnyO/SameO/Unique $\approx$ 67–73\%), and the best Answer (45.26) and General Accuracy (75.86). Since GIS tools require coordinated perceptual grounding, logical inference, and sequential execution, training on the mixed dataset enhances both tool-type alignment (macro F$_1$ across Per./Op./Logic./GIS) and trajectory-level structural coherence. By integrating short and long reasoning traces, the model simultaneously benefits from focused perceptual supervision provided by easy samples and structured multi-step planning signals from hard samples, resulting in improved sequence fidelity under strict order and multiplicity constraints.

\subsection*{Comparison with In-context Learning}
To evaluate the impact of in-context learning, we conduct an ablation study on Qwen3-Instruct-2507 by progressively increasing the number of in-context examples ($k \in \{5, 10, 15\}$) and comparing against the baseline model without demonstrations (BL). 
In-context examples are not appended as a single prompt block, but instead are prepended as a structured conversational history. Specifically, each example is injected as a sequence of user–assistant interaction turns (including intermediate thoughts, tool actions, and observations) before the actual test query. The full message list, therefore, consists of: (1) a system instruction, (2) k prior example dialogues formatted as multi-turn interactions, and (3) the target query conversation. Results are summarized in Table~\ref{tab:abl2}. 

\begin{table*}[h!]
\small
\centering
\caption{Performance comparison of Qwen3-Instruct-2507 under increasing in-context example sizes (k = 5, 10, 15) against the baseline (BL) and OpenEarthAgent. Increasing the number of in-context examples consistently improves performance over the baseline, while OpenEarthAgent achieves the strongest results across all metrics.}
\label{tab:abl2}
\resizebox{\textwidth}{!}{%
\begin{tabular}{lccccccccc}
\hline
\multirow{2}{*}{\textbf{Models}} &
\multicolumn{4}{c}{\textbf{F$_1$ scores}} &
\multicolumn{3}{c}{\textbf{Tool Order}} &
\multicolumn{2}{c}{\textbf{Accuracy}} \\
\cmidrule(lr){2-5}\cmidrule(lr){6-8}\cmidrule(lr){9-10}
&\textbf{Per.} & \textbf{Op.} & \textbf{Logic.} & \textbf{GIS.} &
\textbf{AnyO.} & \textbf{SameO.} & \textbf{Unique} & \textbf{Ans.} &
\textbf{Gen.} \\
\midrule
\rowcolor{OpticalBlue!60}
\multicolumn{10}{c}{\textsc{Qwen3-Instruct-2507}} \\
Baseline (BL) &13.86 & 20.95 & 18.35 & 71.82 & 16.00 & 14.71 & 21.47 & 13.72 & 15.86 \\
BL+ In-context(k=5)  &20.26 & 37.34 & 30.20 & 87.27 & 29.34 & 28.57 & 31.73 & 26.84 & 44.14 \\
BL+ In-context(k=10)  & 24.86 & 46.29 & 34.02 & 86.54 & 30.88 & 30.28 & 33.28 & 28.73 & 51.72 \\
BL+ In-context(k=15)  & 34.58 & 42.10 & 37.80 & 89.80 & 36.44 & 36.10 & 39.44 & 33.06 & 46.90 \\
\rowcolor{OpticalBlue!60}
\multicolumn{10}{c}{\textsc{OpenEarthAgent}} \\
OpenEarthAgent   & \textbf{58.30} & \textbf{56.76} & \textbf{51.18} & \textbf{98.52} & \textbf{67.75} & \textbf{67.24} & \textbf{72.71} & \textbf{45.26} & \textbf{75.86}\\
\hline
\end{tabular}}

\end{table*} 
Adding in-context examples yields substantial improvements across all evaluation dimensions. Even with k=5, performance improves markedly over the baseline, particularly in Operation, GIS F$_1$ scores, and accuracy. Increasing k to 10 further enhances the Logical and Operation F$_1$ scores and yields notable gains in tool-order consistency metrics (AnyO., SameO., and Unique), suggesting that additional demonstrations help the model better internalize structured tool usage patterns.

For k=15, Perception and Logical F$_1$ scores show improvements, and tool-order metrics also consistently increase, indicating improved procedural alignment. However, degradation in Operation and Generation performance between k=10 and k=15 suggests diminishing returns beyond a certain number of demonstrations.
Despite these gains, the model remains substantially below OpenEarthAgent, which outperforms all Qwen variants across every metric. This highlights that while in-context learning significantly enhances tool reasoning and execution consistency, architectural or training-level adaptations, as employed by OpenEarthAgent, are necessary to achieve state-of-the-art performance.

\begin{table*}[b!]
\small
\centering
\caption{Impact of Tool Schema Complexity on Model Reasoning and Tool Usage: Structural exposure (Sch.1→2) improves tool perception and selection; argument-level signatures (Sch.3) enhance interface alignment and GIS reasoning; executable constraints (Sch.4) strengthen logical consistency; and maximal specification with explicit JSON examples (Sch.5) yields the highest overall F$_1$, tool-order correctness, and answer accuracy.}
\label{tab:abl3}
\resizebox{\textwidth}{!}{%
\begin{tabular}{lccccccccc}
\hline
\multirow{2}{*}{\textbf{Tool Schema}}&
\multicolumn{4}{c}{\textbf{F$_1$ scores}} &
\multicolumn{3}{c}{\textbf{Tool Order}} &
\multicolumn{2}{c}{\textbf{Accuracy}} \\
\cmidrule(lr){2-5}\cmidrule(lr){6-8}\cmidrule(lr){9-10}
&\textbf{Per.} & \textbf{Op.} & \textbf{Logic.} & \textbf{GIS.} &
\textbf{AnyO.} & \textbf{SameO.} & \textbf{Unique} & \textbf{Ans.} &
\textbf{Gen.} \\
\midrule
\rowcolor{OpticalBlue!60}
\multicolumn{10}{c}{\textsc{OpenEarthAgent}} \\
Sch.1-Minimal & 19.80 & 8.40 & 32.00 & 58.99 & 25.32 & 24.89 & 26.69 & 8.44 & 17.93 \\
Sch.2-Semantic &43.53 & 27.24 & 35.80 & 64.71 & 26.96 & 26.45 & 31.58 & 9.10 & 25.86 \\
Sch.3-Signature &48.76 & 28.86 & 34.22 & 76.32 & 26.17 & 25.66 & 29.77 & 12.77 &  40.69\\
Sch.4-Executable &47.01 & 27.91 & 40.94 & 76.86 & 26.55 & 25.95 & 29.26& 13.03 & 46.21\\
Sch.5- Maximal  & \textbf{58.30} & \textbf{56.76} & \textbf{51.18} & \textbf{98.52} & \textbf{67.75} & \textbf{67.24} & \textbf{72.71} & \textbf{45.26} & \textbf{75.86}\\
\hline
\end{tabular}}
\end{table*} 

\subsection*{Tool Schema Influence}
To assess the agent's sensitivity to changes in the tool specification format (Table ~\ref{tab:abl3}), we conduct a controlled ablation over five variants ordered from minimal structural exposure to maximal semantic conditioning.

\textbf{Schema 1- Minimal Structure:} This setting adds a thin layer of operational structure; a compact enumeration of all tools and a procedural rule set (e.g., one action per step, Terminate required, tool-order constraints, JSON format). No argument signatures, semantic descriptions, or usage examples are provided.
This schema provides structural benefits for planning while withholding any semantic cues about the tool's functionality. Consequently, the agent must rely on actual tool feedback to infer how each tool behaves.

\textbf{Schema 2- Natural-language semantics:}
Global planning rules and execution constraints remain identical to Schema 1. However, high-level natural-language descriptions of tool functionality are provided along with tool enumeration. The agent gains functional semantic grounding but lacks explicit demonstrations of argument structure, format exemplars, and concrete API usage patterns. This schema isolates the effect of semantic grounding without interface exemplification.

\textbf{Schema 3- Compressed Signature:} The third variant introduces argument level explicitness along with natural-language semantics. Tools are presented as Tool-Aurgument signatures, for example \texttt{Change\allowbreak Detection\allowbreak (text,\allowbreak pre\allowbreak \_image\allowbreak ,post\_image)}. This variant provides only tool-argument information, without any detailed examples. It effectively mimics a code-completion-style API, requiring the agent to infer the tool's behavior from the structure of the signatures.

\textbf{Schema 4- Tool signature + Executable Constraint:}
This configuration exposes the agent to tool signatures similar to Schema 3, along with all global rules. An additional requirement is added for \texttt{command} arguments: they must contain executable Python code defining a \texttt{solution()} function, along with a single example. This operational executability constraint is critical because certain tools cannot operate without a fully defined executable function; for example, \texttt{Solver} uses \texttt{solution()} to perform symbolic reasoning using Sympy, and \texttt{Plot} returns a matplotlib figure generated by code. By enforcing executable commands, the schema ensures that the model produces actionable tool inputs for the critical tool.

\textbf{Schema 5- Maximal Information:}
This represents the upper bound of tool-conditioning information. Each element contributes crucially: purpose-level explanations clarify the tool's goal, argument semantics specify how inputs affect outputs, usage examples demonstrate correct application, and tool rules guide multi-step planning and sequencing. This schema enables the model to reliably understand, plan, and execute complex tool chains. The agent operates under a fully specified tool-use ontology, minimizing ambiguity.

The Table~\ref {tab:abl3} shows that tool-using LLM agents are highly sensitive to how tool knowledge is structured and exposed. Performance differences arise from information exposure, specifically from the extent of semantic grounding and interface clarity.
With minimal structure (Sch. 1), the model relies only on structural cues and feedback, resulting in weak reasoning and the lowest F$_1$ scores, particularly in perception and operation (19.80 and 8.40), indicating that the model struggles to correctly identify which tools to invoke. Introducing natural-language tool descriptions in Schema 2 (Semantic) yields a substantial improvement across F$_1$ metrics. Perception more than doubles (19.80 → 43.53), and operation improves dramatically (8.40 → 27.24). Tool-order scores also rise, suggesting that semantic grounding enables the model to better anticipate more accurate tool use and reasonable action sequences.

Adding explicit argument signatures (Schema 3) improves alignment and task–tool mapping: Perception increases (+5.23), and GIS rises substantially (+11.61), indicating stronger structural grounding. Tool-order metrics remain largely unchanged, suggesting that signature-level exposure enhances invocation precision but not higher-level planning. Generation accuracy increases markedly (+14.83), indicating better abstraction and parameterization when invoking image-generation tools. Introducing executable constraints (Schema 4) primarily enhances reasoning rigor as seen by logical F$_1$ increase (+6.72). Tool-order metrics again show minimal change, indicating that executability strengthens operational validity rather than sequencing strategy.
 
Full specification combines purpose descriptions, argument semantics, executable examples, and planning rules and yields the largest gains: Perception (+11.29), Operation (+28.85), Logical (+10.24), GIS (+21.66), and major improvements in tool-order correctness (AnyO. +41.20; SameO. +41.29; Unique +43.45). These improvements are primarily driven by explicit full JSON tool examples, which provide maximal interface grounding by explicitly instantiating valid invocation patterns. Compared to abstract signatures or natural-language descriptions, these examples define argument schemas, formatting constraints, required fields, quoting conventions, and executable structure. This reduces syntactic ambiguity and malformed calls under strict JSON-only constraints, thereby improving the validity of invocations, the completeness of arguments, and overall schema compliance.

\subsection*{Effect of Tool Order}
This ablation investigates the effects of tool-list ordering and explicit procedural constraints on agent reasoning and execution behavior. We evaluate three configurations: (1) a shuffled tool list with dependency constraints preserved, (2) an optimized tool list without explicit dependency rules, and (3) an optimized tool list with explicit dependency rules enforcing valid tool calling preconditions. The first setting disrupts positional priors by randomizing the presentation of tools. The second retains a structured presentation but removes hard-coded ordering constraints, requiring the model to infer dependencies implicitly. The third combines a structured tool presentation with explicit procedural guidance.

The results (Table~\ref{tab:abl4})demonstrate that tool ordering has a substantial impact on reasoning performance and execution behavior. Progressing from shuffled tool order to optimized ordering and further to optimized ordering with an explicit order rule yields consistent improvements in F$_1$ scores, particularly for Perception $(+6.80, +1.85)$, Logic $(+3.93, +2.68)$, and GIS $(+2.98, +1.06)$. These gains indicate that structured sequencing enhances information retrieval, multi-step dependency handling, and spatial reasoning. Operational performance peaks under optimized ordering without the rule (59.54), suggesting that strict ordering constraints may marginally reduce operational flexibility.

\begin{table*}[hbt!]
\small
\centering
\caption{Impact of Tool Order on Model Reasoning and Tool Usage: Comparing shuffled order, optimized order without rules, and optimized order with enforced preconditions shows consistent gains in reasoning F$_1$, tool-order correctness, and answer Acc., while Generation Acc. remains stable.}

\label{tab:abl4}
\resizebox{\textwidth}{!}{%
\begin{tabular}{lccccccccc}
\hline
\multirow{2}{*}{\textbf{Tool Schema}}&
\multicolumn{4}{c}{\textbf{F$_1$ scores}} &
\multicolumn{3}{c}{\textbf{Tool Order}} &
\multicolumn{2}{c}{\textbf{Accuracy}} \\
\cmidrule(lr){2-5}\cmidrule(lr){6-8}\cmidrule(lr){9-10}
&\textbf{Per.} & \textbf{Op.} & \textbf{Logic.} & \textbf{GIS.} &
\textbf{AnyO.} & \textbf{SameO.} & \textbf{Unique} & \textbf{Ans.} &
\textbf{Gen.} \\
\midrule
\rowcolor{OpticalBlue!60}
\multicolumn{10}{c}{\textsc{OpenEarthAgent}} \\
Shuffled Tool Order & 49.65 & 58.97 & 44.57 & 94.48 & 63.22 & 62.36 & 67.92 & 43.35 &  \textbf{75.93}\\
Optimized Order w/o Order Rule &56.45 & \textbf{59.54} & 48.50 & 97.46 & 65.86 & 64.84 & 71.26 & 45.14 & 75.17\\
Optimized Order w/ Order Rule  & \textbf{58.30} & 56.76 & \textbf{51.18} & \textbf{98.52} & \textbf{67.75} & \textbf{67.24} & \textbf{72.71} & \textbf{45.26} & 75.86\\
\hline
\end{tabular}}
\end{table*} 
Tool-order correctness metrics (AnyO., SameO., Unique) improve monotonically across settings, reflecting increased consistency in tool invocation patterns. Answer accuracy also improves modestly (43.35→45.26), whereas generation accuracy remains stable, indicating that ordering primarily enhances execution reliability and in-distribution reasoning rather than expanding overall generative capacity.

\subsection*{Impact of High-Level Planning}
We compare OpenEarthAgent under two configurations: (1) without high-level planning, where the agent directly invokes tools with only step-level planning, and(2) with high-level planning, where the agent generates an explicit reasoning plan prior to tool execution. Both variants are trained and evaluated on the same dataset, using identical tools, supervision, and evaluation protocols.

The results (Table ~\ref{tab:High_level_plan}) show that incorporating a high-level planning step leads to consistent improvements. Logical reasoning F1 increases by +3.57, reflecting improved multi-step consistency. Tool sequencing also improves, with Any-order accuracy +4.36 and Same-order accuracy +4.37, indicating more coherent and reliably structured tool execution under explicit planning. Moreover, image generation accuracy increases from 65.52 to 75.86 (+10.34), indicating that explicit high-level planning significantly improves the accuracy of deciding when to generate an image in response to a query.

\begin{table}[b!]
\caption{Effect of High-Level Planning: Explicit high-level planning improves logical reasoning, tool sequencing consistency, and image generation decision accuracy, leading to stronger structured execution and overall performance.}
\label{tab:High_level_plan}
\small
\centering
\resizebox{\textwidth}{!}{%
\begin{tabular}{lccccccccc}
\hline
\multirow{2}{*}{\textbf{Setting}}&
\multicolumn{4}{c}{\textbf{F$_1$ scores}} &
\multicolumn{3}{c}{\textbf{Tool Order}} &
\multicolumn{2}{c}{\textbf{Accuracy}} \\
\cmidrule(lr){2-5}\cmidrule(lr){6-8}\cmidrule(lr){9-10}
&
\textbf{Per.} &
\textbf{Op.} &
\textbf{Logic.} &
\textbf{GIS.} &
\textbf{AnyO.} &
\textbf{SameO.} &
\textbf{Unique} &
\textbf{Ans.} &
\textbf{Gen.} \\
\midrule
\rowcolor{OpticalBlue!60}
\multicolumn{10}{c}{\textsc{OpenEarthAgent}} \\
w/o High-level Plan  & 57.74 & \textbf{58.82} & 47.61 & 97.11 & 63.39 & 62.87 & 71.77 & 44.50 & 65.52 \\
w/ High-level Plan & \textbf{58.30} & 56.76 & \textbf{51.18} & \textbf{98.52} & \textbf{67.75} & \textbf{67.24} & \textbf{72.71} & \textbf{45.26} & \textbf{75.86}\\
\hline
\end{tabular}}
\end{table}

\section{Model Selection}
\stepcounter{secnumber}
\label{sec:modelselection}

To determine how underlying model capacity affects tool-use accuracy and multi-step reasoning, we compare two model backbones trained under identical data, hyperparameter, and tool-conditioning settings: Qwen3-4B-Instruct-2507 and Qwen2.5-7B-Instruct.
We train on 25\% of the training corpus and reserve an internal 8\% split from it as a held-out validation set used only for model selection, entirely separate from the benchmark test set.

As summarized in Table~\ref{tab:model_sel}, the smaller Qwen3-4B-Instruct-2507 outperforms the 7B counterpart across nearly all metrics, particularly in logical reasoning (Logic F$_1$: +2.79) and tool-order consistency (AnyO: +7.0). This indicates that architectural refinements and improved alignment in the Qwen3 series enhance structured reasoning efficiency even with fewer parameters.  
Both variants exhibit near-saturated GIS and operational F$_1$ scores, confirming that tool-augmented reasoning generalizes robustly across these domains. The marginal differences in answer and generation accuracy further suggest that reasoning quality is independent of model scale.  
Overall, Qwen3-4B-Instruct-2507 delivers a superior balance between model size and reasoning reliability, making it a strong candidate for a baseline model for scalable geospatial applications.

\begin{table}[h!]
\caption{Comparison of base model performance across reasoning and tool-usage metrics. Qwen3-4B-Instruct-2507 slightly outperforms Qwen2.5-7B-Instruct. Results highlight that improved architectural alignment in the Qwen3 series yields 
more efficient multi-step reasoning despite a smaller model size.}
\label{tab:model_sel}
\small
\centering
\resizebox{\textwidth}{!}{%
\begin{tabular}{lccccccccc}
\hline
\multirow{2}{*}{\textbf{Base Model}}&
\multicolumn{4}{c}{\textbf{F$_1$ scores}} &
\multicolumn{3}{c}{\textbf{Tool Order}} &
\multicolumn{2}{c}{\textbf{Accuracy}} \\
\cmidrule(lr){2-5}\cmidrule(lr){6-8}\cmidrule(lr){9-10}
&
\textbf{Per.} &
\textbf{Op.} &
\textbf{Logic.} &
\textbf{GIS.} &
\textbf{AnyO.} &
\textbf{SameO.} &
\textbf{Unique} &
\textbf{Ans.} &
\textbf{Gen.} \\
\midrule
\rowcolor{OpticalBlue!60}
\multicolumn{10}{c}{\textsc{OpenEarthAgent}} \\
Qwen2.5-7B-Instruct & 96.72 & 97.41 & 88.60 & 99.53 & 74.80 & 74.80 & 96.00 & 23.19 & 89.67 \\
Qwen3-4B-Instruct-2507 & \textbf{97.82} & \textbf{97.70} & \textbf{91.39} & 99.53 & \textbf{81.80} & \textbf{81.60} & \textbf{97.50} & \textbf{23.47} & \textbf{90.76} \\
\hline
\end{tabular}}
\end{table}

\section{Error Analysis}
\stepcounter{secnumber}
\label{sec:erroranalysis}

\begin{figure}[h] 
  \centering
  \includegraphics[width=0.9\columnwidth]{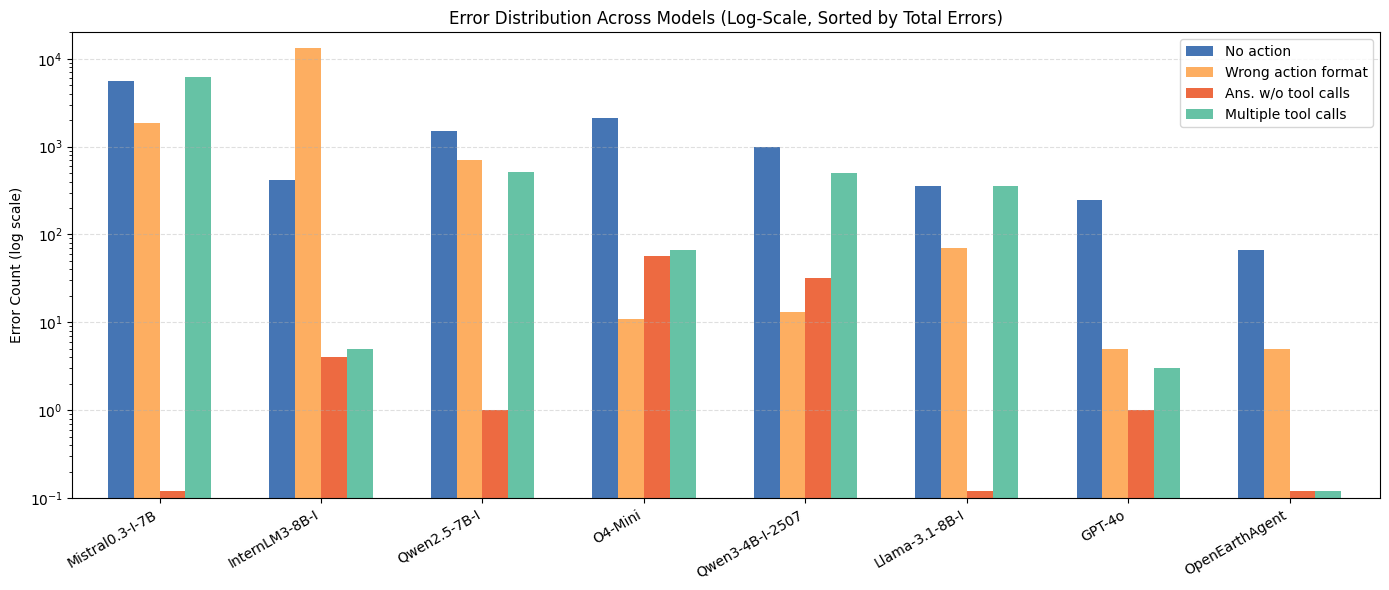}
  \caption{Log-scale error distribution across models for four error types: No action, Wrong action format, Answer reached without a single tool call, and Multiple tool calls in a single action. Frontier models show fewer errors overall, while the fine-tuned OpenEarthAgent achieves the lowest total error count.}
  \label{fig:error_breakdown}
\end{figure}

We conduct a fine-grained analysis of model failures to better understand where tool-augmented reasoning breaks down. 
Fig.~\ref{fig:error_breakdown} shows errors across 4 types, and reports their frequency across some evaluated models.

Syntax errors occur when the model fails to comply with the prescribed output syntax, such as missing required actions(No Action) or producing malformed JSON structures (Wrong action format). 
Each conversation allows a single ``thought-only'' turn for initial planning, but all subsequent turns are expected to include at least one valid action. Open-source models exhibit notably higher rates of such violations, particularly \texttt{Mistral0.3-7B-Instruct} and \texttt{InternLM3-8B-Instruct}, which frequently omit or incorrectly format tool calls. 
In contrast, \texttt{GPT-4o} shows substantially fewer syntax-related failures, indicating stronger adherence to structured output specifications.
\textsc{OpenEarthAgent} exhibits the fewest syntax errors overall, suggesting that fine-tuned tool schema conditioning substantially improves output validity.

The reasoning errors are failures in multi-step logic and tool selection, including queries that reach an answer without a single tool call (Ans. w/o tool call) and redundant tool calls in a single step (Multiple tool calls). 
``Answer w/o tool call'' errors are rare across models, showing that most models correctly trigger tools when needed, except \texttt{Qwen3-4B-Instruct-2507} and \texttt{o4-mini}, which omit calls more often. 
Conversely, ``Multiple tool calls'' are more prevalent, particularly in \texttt{Mistral0.3-7B-Instruct}, indicating over-generation and weaker action control. 
Frontier model (\texttt{GPT-4o}) maintains tighter regulation, while \textsc{OpenEarthAgent} achieves the best balance, invoking tools only when required and avoiding both omissions and redundancies.

\begin{figure}[t!]
  \centering
  \begin{minipage}[t]{0.41\columnwidth}
    \centering
    \includegraphics[width=\linewidth]{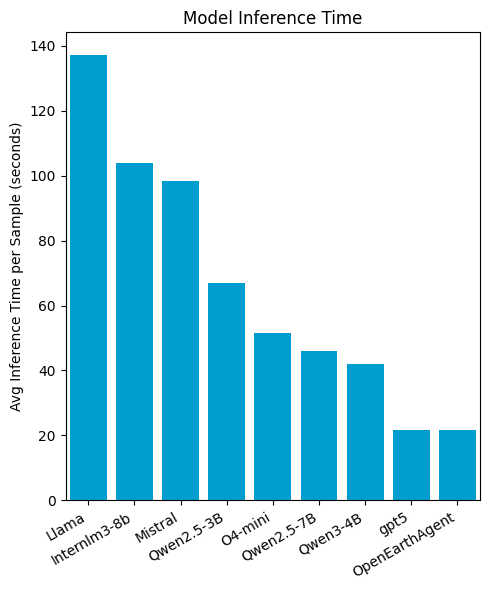}
  \end{minipage}
  \begin{minipage}[t]{0.57\columnwidth}
    \centering
    \includegraphics[width=\linewidth]{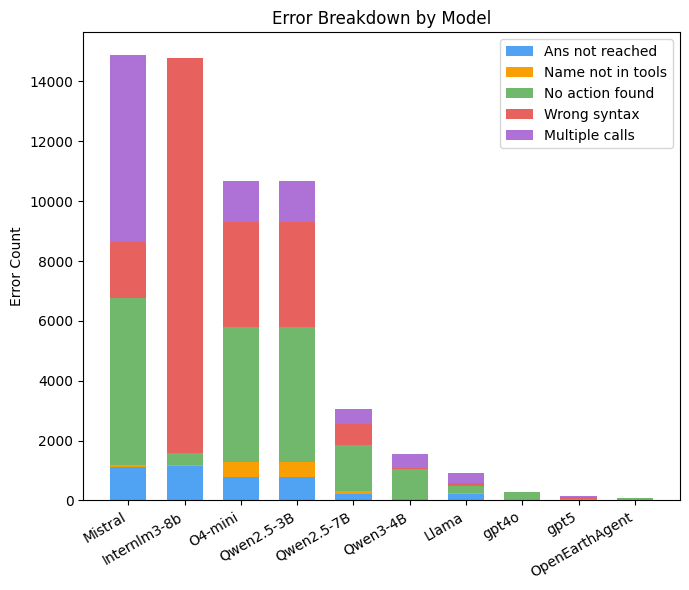}
  \end{minipage}
  \vspace{-0.6em}
  \caption{Comparative model performance:\textbf{(Left)} Inference time efficiency across architectures. \textbf{(Right)} Error distribution by type, revealing tool-specific failure modes per model.}
  \vspace{-0.7em}
  \label{fig:side_by_side}
\end{figure}

Fig.~\ref{fig:side_by_side} compares inference latency and error patterns across models. 
\texttt{Llama\allowbreak 3.1-8B}, \texttt{InternLM3-8B} and \texttt{Mistralv0.3-7B} show high inference times (98--137~s/sample), while \texttt{GPT-5} 
and \texttt{Open\allowbreak Earth\allowbreak Agent} achieve $\sim$21.66 and $\sim$21.67 s/sample respectively. Error analysis reveals that tool-related failures dominate, primarily syntax errors and failures to call required tools, indicating challenges in structured tool invocation. \texttt{GPT-4o}, \texttt{GPT-5}, and \texttt{OpenEarthAgent} show substantially fewer 
errors across all categories. Overall, \texttt{OpenEarthAgent} and frontier proprietary models achieve lower inference latency, better tool-use reliability, and fewer execution errors, demonstrating superior overall agent performance.

\section{Context-Specific Performance Assessment}
\stepcounter{secnumber}
\label{sec:contanalysis}

We further analyze model performance by stratifying the OpenEarthAgent Benchmark across domain-specific categories, including urban planning, environmental monitoring, recreation, industrial infrastructure, disaster assessment, aviation, transportation, and others. This breakdown enables fine-grained evaluation of contextual robustness. Domain-specific difficulty in remote sensing arises from variations in spatial scale, temporal dynamics, spectral ambiguity, and semantic complexity. Structured domains such as urban planning and environmental monitoring typically exhibit stable, large-scale spatial patterns and consistent spectral signatures, making them comparatively easier to analyze. In contrast, disaster assessment involves abrupt temporal changes, heterogeneous damage patterns, and noise, increasing change-detection complexity. Industrial, aviation, and transportation tasks often require fine-grained recognition of small or sparse objects under varying resolutions and viewpoints, demanding stronger contextual and geometric reasoning.

\begin{figure}[t!] 
  \centering
  \includegraphics[width=\columnwidth]{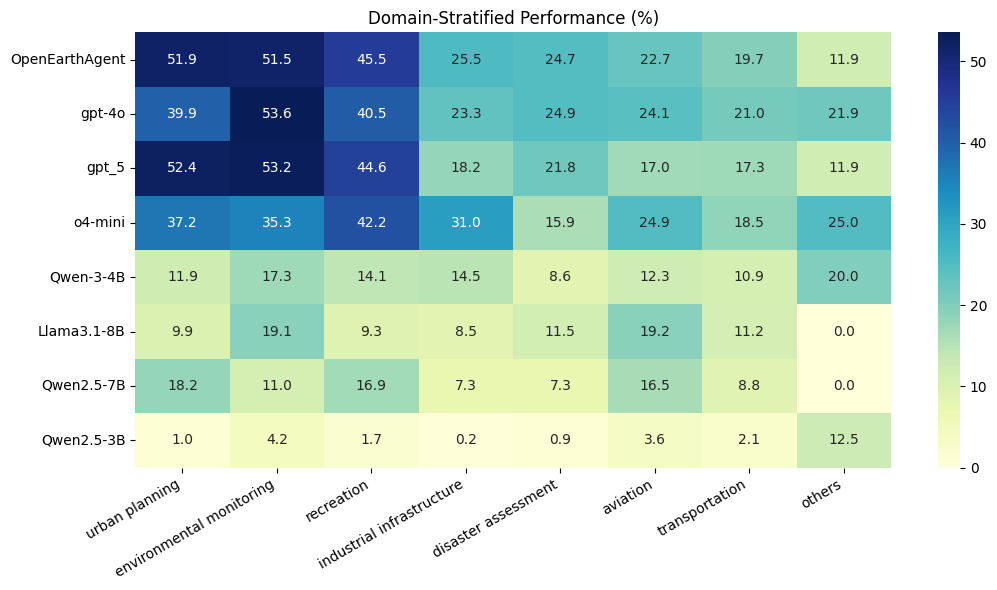}
  \caption{Domain-stratified performance on the OpenEarthAgent Benchmark. The heatmap shows accuracy across eight geospatial domains. OpenEarthAgent delivers strong, consistent results, particularly in urban planning and environmental monitoring, while remaining competitive in infrastructure and disaster-related tasks.}
  \label{fig:domain}
\end{figure}

As shown in Figure~\ref{fig:domain}, OpenEarthAgent demonstrates strong and consistent performance across high-frequency geospatial domains, particularly in urban planning, environmental monitoring, and recreation, where it matches or surpasses frontier models such as GPT-4o and GPT-5. Performance remains competitive in infrastructure and disaster-related tasks, indicating effective generalization to operational geospatial reasoning scenarios. In comparison, open-source baselines exhibit substantially greater performance variance and lower accuracy in complex reasoning tasks.

\section{Qualitative Results}
\stepcounter{secnumber}
\label{sec:qualitative}

\begin{figure*}[hb!]
  \centering
  \includegraphics[width=\linewidth]{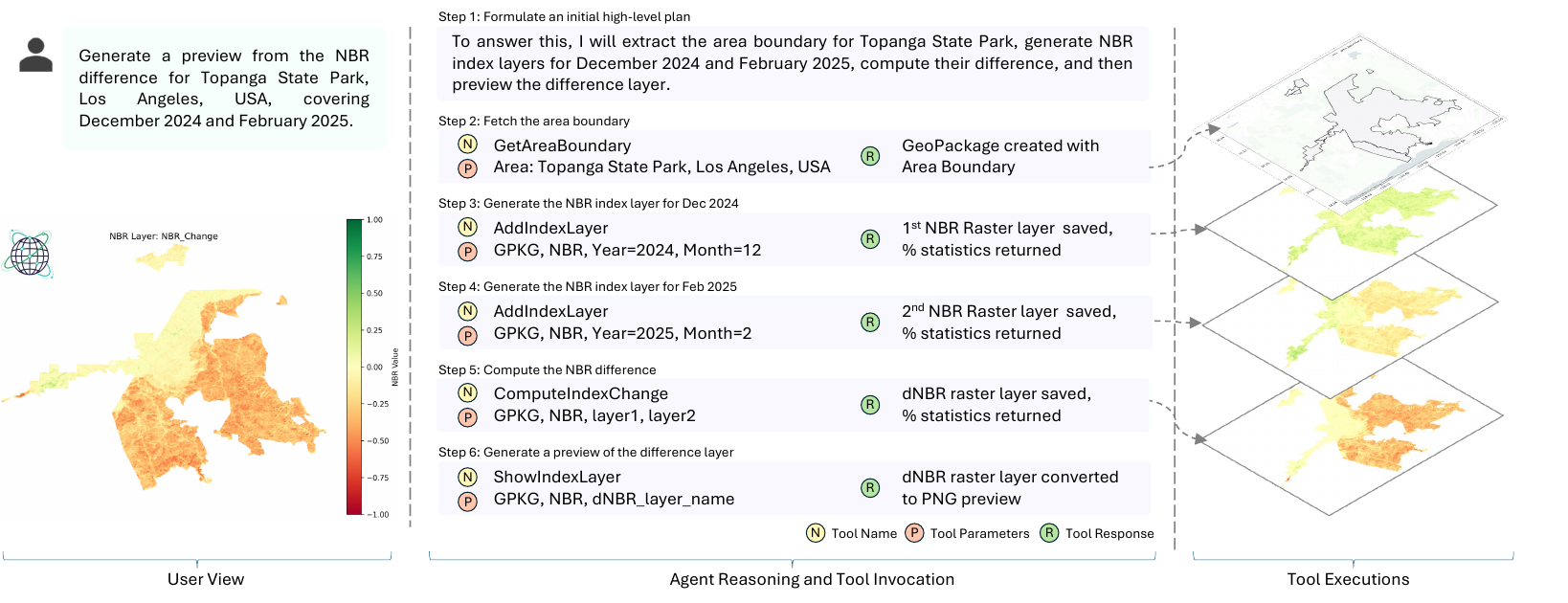}
  \caption{Example illustrates the zero-shot geospatial reasoning capabilities of OpenEarthAgent on a real-world environmental task by generating a Normalized Burn Ratio (NBR) difference map for a given region. The NBR difference map shows pronounced negative values (red) in southern Topanga, revealing burn scars and vegetation loss, while northern areas remain stable (yellow), showing the post-fire effects from early 2025 events.} 
  \label{fig:sample1nbr}
\end{figure*}

Fig.~\ref{fig:sample1nbr} presents a zero-shot example demonstrating the geospatial reasoning capability of \text{OpenEarthAgent}. Given a natural language instruction, the agent autonomously composes a complete multi-step workflow. The agent first formulates a plan to extract the park boundary, compute NBR layers, evaluate their difference, and visualize the result. It sequentially invokes the tools \texttt{Get\allowbreak Area\allowbreak Boundary}, \texttt{Add\allowbreak Index\allowbreak Layer} (for December 2024 and February 2025), \texttt{Compute\allowbreak Index\allowbreak Change}, and \texttt{Show\allowbreak IndexLayer}, generating a differenced NBR (dNBR) layer that quantifies vegetation loss and recovery. The reasoning trace(middle) of Fig.~\ref{fig:sample1nbr} illustrates coherent tool grounding and correct argument formatting across the entire chain. 

\begin{figure*}[hb!]
  \centering
  \includegraphics[width=0.8\linewidth]{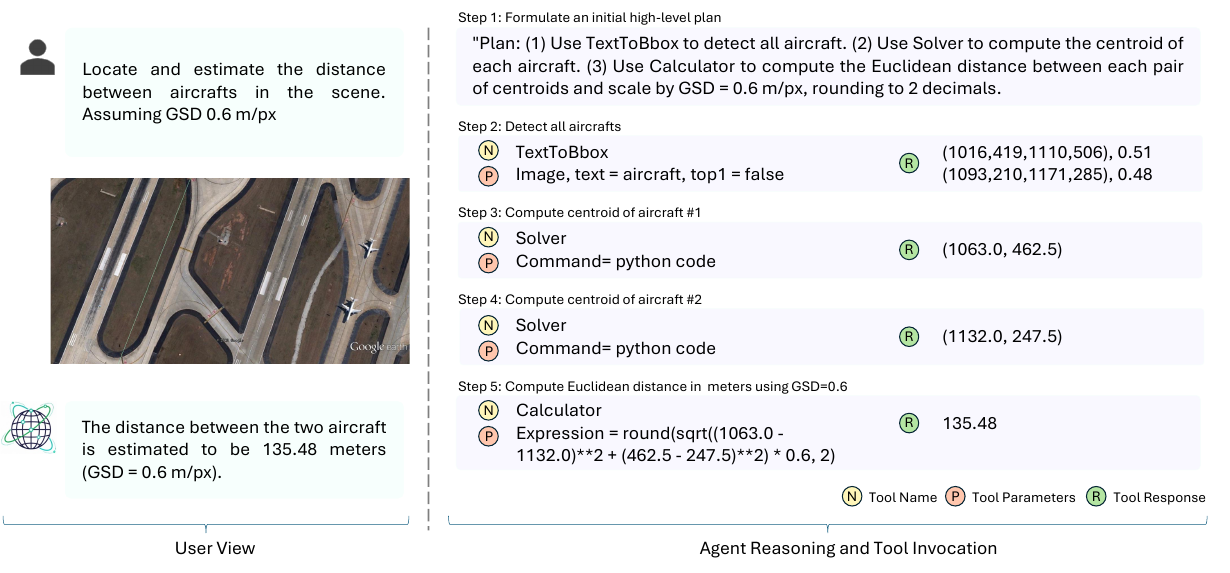}
  \caption{Example demonstrating \textsc{OpenEarthAgent}’s capability for metric-scale geospatial reasoning from imagery, estimating an airplane-to-airplane distance using centroid detection and GSD scaling.} 
  \label{fig:sample2}
\end{figure*}

\begin{figure*}[hbt!]
  \centering
  \includegraphics[width=0.8\linewidth]{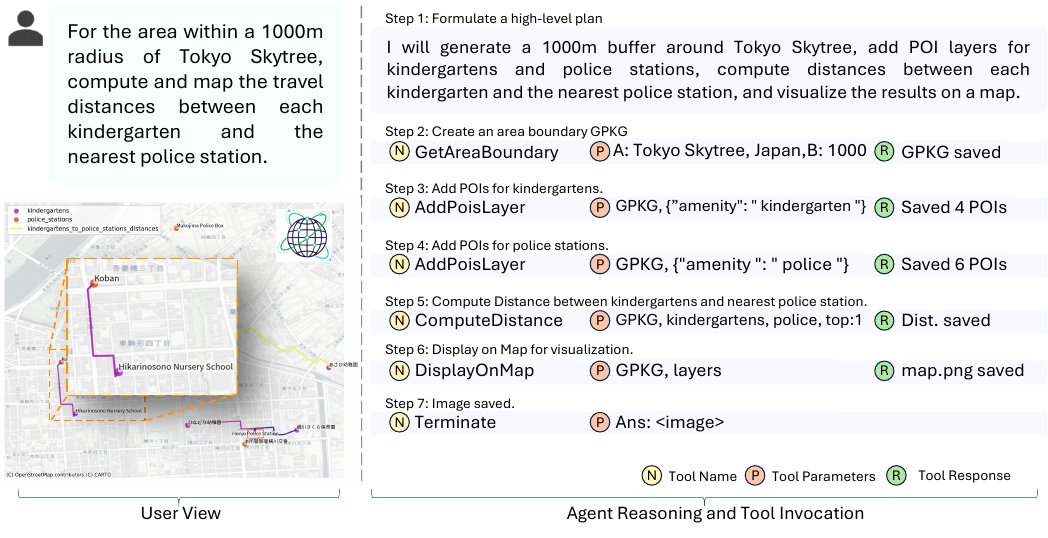}
  \caption{Example demonstrating OpenEarthAgent’s road-network distance computation and spatial analysis within a 1000 m buffer of a specified area. The visualization highlights the detailed map output generated through sequential tool orchestration.} 
  \label{fig:sample3}
\end{figure*}

Fig.~\ref{fig:sample2} illustrates an example of \textsc{OpenEarthAgent} performing metric-scale geospatial reasoning. Given a natural-language query about the distance between two vehicles with a GSD of 0.6, the agent composes a workflow: detecting vehicles via \texttt{TextToBbox}, computing centroids with \texttt{Solver}, and calculating Euclidean distance using \texttt{Calculator}. The reasoning trace shows coherent tool use and precise cross-modal computation. In Fig.~\ref{fig:sample3}, the agent extracts the buffered area, adds POI layers, computes nearest-neighbor road distances from each kindergarten to the closest police station, and visualizes the POIs and distance links on a base map. This demonstrates the agent’s ability to integrate spatial query and interpretable visualization within a unified reasoning pipeline. 

\section{Additional Tools Details}
\stepcounter{secnumber}
\label{sec:tooldescription}

\textbf{Tools I/O Structure:} The tools used in OpenEarthAgent span a wide range of perceptual, geometric, and analytical operations. 
They include basic utilities such as drawing boxes, adding text, OCR, and solvers, etc., alongside higher-level tools for object detection, region description, change detection, and pixel-level segmentation. 
GIS functions such as retrieving area boundaries, adding POI layers, and computing distances, etc., provide the spatial backbone needed for grounded reasoning. 
Spectral tools further extend this capability by generating and comparing index layers such as NDVI or NDBI. 
Each tool follows a clear schema with defined inputs and outputs (see Table~\ref{tab:tool-table}), allowing the agent to compose them into transparent, multi-step reasoning chains across imagery, metadata, and geospatial layers.

\noindent\textbf{Tool Implementations:} OpenEarthAgent integrates diverse tools to support different stages of the reasoning and analysis pipeline. For remote sensing change detection, we employ \texttt{TEOChat}~\cite{irvin2024teochat}, while \texttt{LAE-DINO}~\cite{pan2025locate} is used for object detection and localization in Earth observation imagery. Geospatial index computation and environmental analysis are implemented through \texttt{Google Earth Engine (GEE)} and \texttt{geemap}, providing access to a wide range of remote sensing products and spectral indices. For spatial querying, routing, and network-based analysis, we utilize \texttt{OSMnx} and \texttt{NetworkX}, enabling operations over road networks and OpenStreetMap data. Image understanding and visual scene description are handled by \texttt{Qwen3-VL-7B}~\cite{bai2025qwen3}. Together, these tools provide the domain-specific capabilities required for complex geospatial reasoning tasks. During evaluation, the complete tool stack requires approximately \texttt{45 GB} of runtime memory, with minimal timeout occurrence.

\section{Evaluation Prompt}
\stepcounter{secnumber}
\label{sec:eval_prompt_sec}

To evaluate the geospatial agent’s final output, we adopt a structured Answer Accuracy assessment framework, illustrated in Fig. ~\ref{fig:eval_prompt}. The evaluation protocol measures the alignment between a model’s predicted answer and the corresponding ground truth using a normalized score [0,1]. The framework distinguishes between numerical and non-numerical outputs. Quantitative responses are evaluated using a $\pm$10\% tolerance threshold, whereas spatial outputs use Intersection over Union (IoU) thresholds. For descriptive or categorical answers, evaluation is based on semantic equivalence, considering paraphrasing, partial correctness, and contradictions. The protocol further defines strict handling of missing, fabricated, or non-responsive outputs, assigning zero scores where required elements are absent or invalid. To standardize assessment, the framework enforces a constrained JSON output format consisting of a numerical score and a concise justification.


\begin{table}[H]
\centering
\scriptsize
\caption{List of available tools with brief descriptions, inputs, and outputs.}
\label{tab:tool-table}
\setlength{\tabcolsep}{2.5pt}      
\renewcommand{\arraystretch}{1.1}   
\rowcolors{2}{gray!15}{white}       
\begin{adjustbox}{width=\textwidth} 
\begin{tabularx}{\textwidth}{%
    L{2.5cm}
    L{3cm}
    L{3cm}
    L{3cm}
}
\toprule
\textbf{Tool Name} & \textbf{Description} & \textbf{Input} & \textbf{Output} \\
\midrule
Calculator & Evaluates math expressions using Python. & Math expression (Python syntax) & Numeric result as text \\
OCR & Extract text from an image. & Image & Text with bounding boxes \\
DrawBox & Draws a box on an image. & Image; box coordinates; optional label & Image with drawn box \\
AddText & Adds text to an image. & Image; text; position; optional color & Image with text overlay \\
GoogleSearch & Returns search results from Google. & Query string; optional number of results & Search result text \\
Plot & Plots data using Python code. & Python code defining \texttt{solution()} & Generated plot image \\
Solver & Solves equations using \texttt{sympy}. & Python code defining \texttt{solution()} and using \texttt{sympy} & Equation solution as string \\
TextToBbox & Finds object based on description. & Image; object description & Bounding box coordinates \\
ImageDescription & Generates a caption for an image. & Image & Text description \\
RegionAttribute\,-\, Description & Describes an attribute in a region or image. & Image; attribute; optional bounding box & Text description of attribute \\
CountGivenObject & Counts objects in an image. & Image; object name; optional bounding box & Integer count \\
ChangeDetection & Describes changes between two images. & Text query; pre- and post-images & Description of changes \\
SegmentObject\,-\, Pixels & Segments objects, and count pixels. & Image; object name; optional flag & Pixel count(s) \\
ObjectDetection & Detects objects in the image. & Image & Labels; boxes; scores \\
GetAreaBoundary & Gets the area boundary as a GeoPackage. & Place name or bounding box; optional buffer & GeoPackage with boundary \\
AddPoisLayer & Adds POIs to a GeoPackage. & GeoPackage; query; layer name & Confirmation with POI count \\
ComputeDistance & Computes distance between layers. & GeoPackage; source layer; target layer & Summary text and layer name \\
DisplayOnMap & Renders the map image from layers. & GeoPackage; layer names & PNG image path \\
AddIndexLayer & Computes spectral index layer. & GeoPackage; index type; year; name & Saved layer and class statistics \\
ComputeIndex\,-\, Change & Computes change between index layers. & GeoPackage; index type; layer names & Saved change layer and statistics \\
ShowIndexLayer & Generates a preview of the raster index. & GeoPackage; index type; layer name & Preview image path \\
GetBboxFrom\,-\, Geotiff & Extracts an area bounding box (W,S,E,N) from a GeoTIFF. & GeoTIFF & GeoPackage with boundary \\
DisplayOnGeotiff & Renders GeoPackage layers (with feature names) over a given GeoTIFF. & GeoPackage; layers; GeoTIFF & Rendered GeoTIFF \\
Terminate & Ends task and returns the final answer. & Final answer text & None \\
\bottomrule
\end{tabularx}
\end{adjustbox}
\end{table}

\subsection*{Open-Source LLM Judge: A Comparative Evaluation}
\label{sec:opensource_judge}
\begin{figure}[hbt!]
  \centering
  \includegraphics[width=0.7\linewidth]{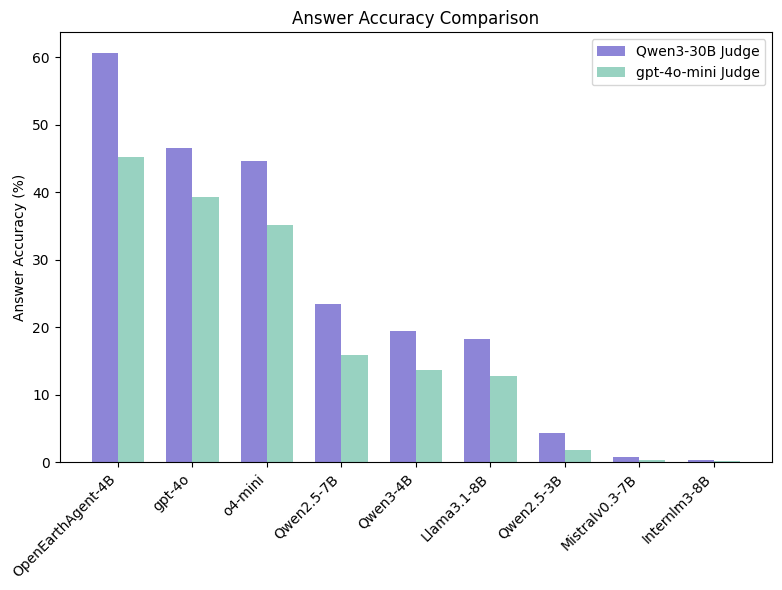}
  \caption{Evaluation comparison for Geospatial Agent Answer Accuracy Assessment. Although Qwen3-30B assigns slightly higher absolute scores, model rankings remain consistent, demonstrating the robustness of the evaluation framework.}
  \label{fig:llm_judge}
\end{figure}

\begin{figure}[hbt!]
  \centering
  \includegraphics[width=\linewidth]{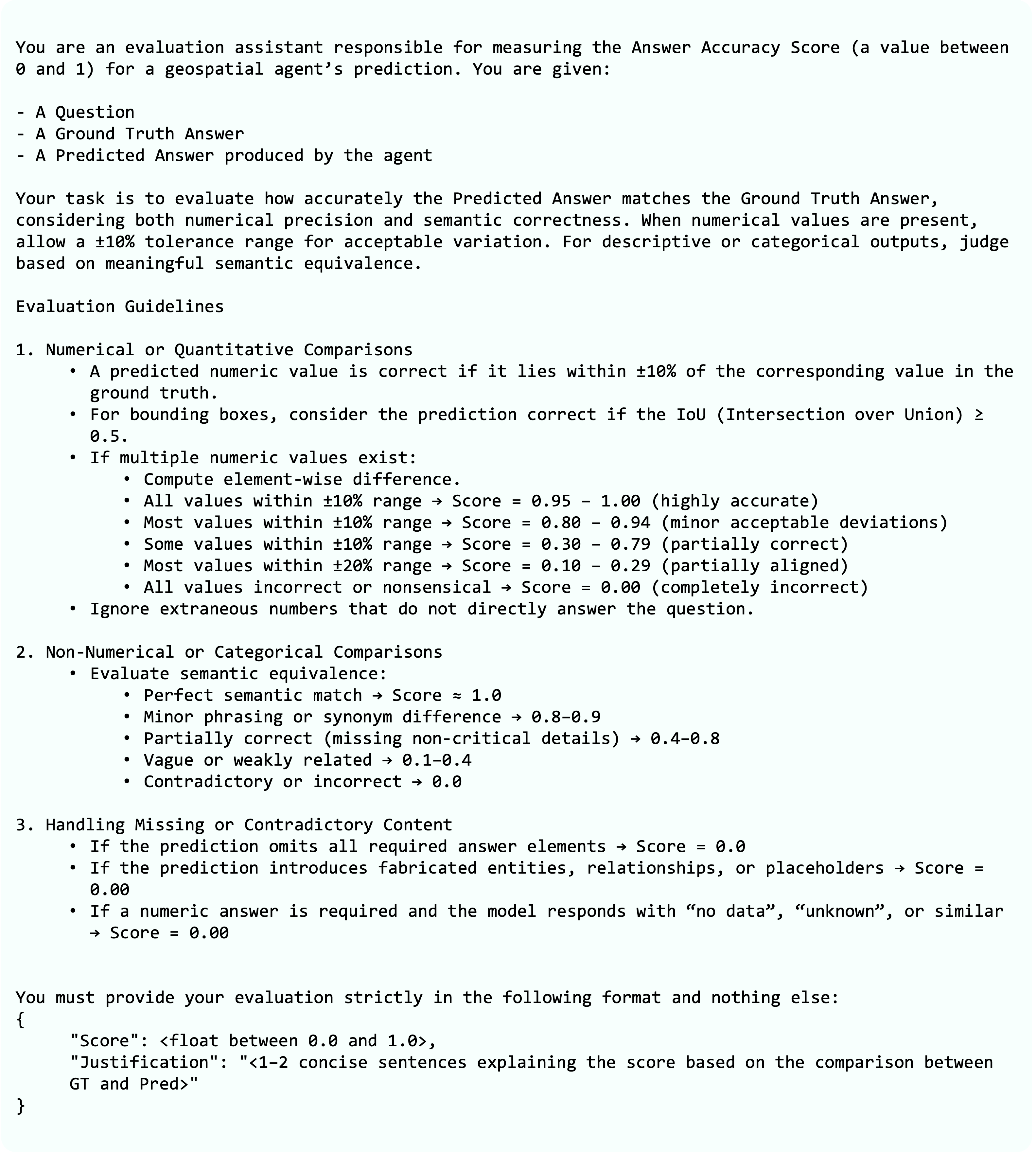}
  \caption{Evaluation Prompt for Geospatial Agent Answer Accuracy Assessment. The figure shows the structured prompt used to compute an Answer Accuracy Score (0–1) for a geospatial agent’s prediction. It defines criteria for numerical tolerance, semantic equivalence, error handling, and required JSON output format for standardized evaluation.}
  \label{fig:eval_prompt}
\end{figure}

To examine the robustness of our evaluation framework, we replace the original proprietary judge with an open-source model, Qwen3-30B, while keeping the evaluation prompt, scoring criteria, and JSON-constrained output format identical (Fig.~\ref{fig:eval_prompt}). This controlled substitution enables a direct cross-judge consistency analysis. As shown in Fig. ~\ref{fig:llm_judge}, Qwen3-30B systematically assigns slightly higher absolute Answer Accuracy scores across models. Despite this positive bias in absolute scoring, the relative ranking of models remains preserved, with stronger agents consistently outperforming weaker baselines under both judges. These findings indicate that while Qwen3-30B tends toward more lenient scoring, the structured evaluation prompt ensures consistent comparative trends, demonstrating the robustness of the proposed Answer Accuracy framework across different LLM judges.

\section{Detailed Description of Dataset-Curation Components}
\stepcounter{secnumber}
\label{sec:dataset_curation_components}

This appendix provides additional implementation details for the data-curation pipeline used to construct the OpenEarthAgent corpus. The goal of the pipeline is to transform heterogeneous earth-observation sources into a unified set of multimodal samples, natural-language queries, and executable reasoning trajectories. Each component in Fig. 2 performs a specific normalization, filtering, or validation step to ensure that the final samples are spatially grounded, tool-compatible, and suitable for supervised trajectory learning.

\subsection*{RGB-SAR Data Ingestion}

The RGB-SAR data ingestion module aggregates candidate samples from multiple open-access datasets containing optical RGB imagery and synthetic aperture radar (SAR) imagery. These datasets provide complementary spatial resolutions, sensor characteristics, and annotation types, including bounding boxes, category labels, segmentation masks, pixel counts, and geospatial metadata. We first download all datasets and parse their native metadata, image paths, annotation files, and sensor-specific attributes. Since the sources differ in file organization, coordinate availability, class taxonomies, and annotation style, this stage only performs source-level extraction and indexing. The output is a pool of candidate RGB and SAR samples that can be further filtered according to the requirements of each reasoning category.

\subsection*{Sample Sufficiency Filter}

Different reasoning tasks require different minimum annotation conditions. For example, counting and comparison queries require a sufficient number of visible objects, proximity and distance questions require at least two relevant entities, direction or attribute questions require localized objects with valid boxes, and change-detection questions require paired pre- and post-event inputs. The sample sufficiency filter enforces these task-dependent constraints before question generation. Given the parsed annotations for each sample, it assigns the sample to one or more predefined reasoning categories only if the required number and type of objects are present. Samples with insufficient objects, ambiguous labels, missing boxes, invalid masks, or incomplete paired inputs are discarded at this stage. This prevents the question synthesizer from generating queries that cannot be answered reliably from the available annotations.

\subsection*{Annotation Harmonizer}

The annotation harmonizer converts heterogeneous annotation formats into a unified representation used by the downstream tool registry. Source datasets provide annotations in different forms, such as horizontal bounding boxes, oriented boxes, segmentation masks, pixel counts, category IDs, and dataset-specific label names. We map these formats into a common JSON-compatible schema containing image identifiers, object categories, bounding-box coordinates, masks where available, spatial metadata, and task-relevant attributes. Label names are normalized across datasets to reduce semantic duplication, and missing fields are filled when they can be derived deterministically from existing annotations. For samples that require additional object attributes not provided by the original dataset, we add manual or semi-automatic annotations. This harmonization step ensures that all RGB and SAR samples can be used by the same perception tools, including object localization, counting, region-attribute description, and visual grounding.

\subsection*{Area Candidate Generator}

The GIS branch begins with an area candidate generator that identifies valid geographic regions for spatial reasoning tasks. Each candidate area is represented by a place name, boundary, bounding box, or GeoPackage-compatible geometry. Candidate regions are selected to support GIS operations such as point-of-interest retrieval, distance computation, buffering, area estimation, and map visualization. The generator also considers whether the region is likely to contain useful infrastructure, transportation, environmental, or urban entities. For queries requiring local accessibility or spatial relationships, buffer distances are added around the target area to create an appropriate region of interest. The output of this module is a set of candidate geographic areas that can be checked for available POI layers and converted into executable geospatial inputs.

\subsection*{Available POI Layers}

For each candidate area, we identify point-of-interest layers that can support meaningful GIS reasoning. These layers are derived from open geospatial sources such as OpenStreetMap and include categories such as hospitals, doctors, schools, colleges, parks, museums, malls, roads, transport nodes, industrial sites, and other urban or infrastructure entities. Each layer is stored with its corresponding tag structure, geometry type, and semantic category. The purpose of this component is to expose the question-generation pipeline to only those POI classes that are compatible with the available tools and likely to produce executable spatial queries.

\subsection*{POI Existence Filter}

The POI existence filter verifies whether a candidate area contains the required POI categories before query synthesis. For example, a distance query between doctors and colleges is only retained if both POI layers exist in the selected region. Similarly, counting, proximity, and accessibility queries require that the requested entities are present in sufficient numbers. This module executes lightweight geospatial checks over the candidate GeoPackage or boundary and discards areas where relevant POIs are missing, too sparse, duplicated, or geometrically invalid. This avoids generating natural-language questions whose tool execution would return empty or uninformative outputs.

\subsection*{Index/Event Miner}

The index-based branch constructs samples around physically meaningful spectral or temporal events. We mine candidate regions and time windows using spectral indicators such as NDVI, NBR, and NDBI, which capture vegetation condition, burn severity, built-up structure, and related environmental changes. The module searches for regions where index values exhibit interpretable temporal patterns, such as vegetation loss, burn damage, regrowth, flooding, or urban expansion. Each candidate event is associated with the required area boundary, date range, index type, and raster source. This produces index-based samples that can support reasoning over environmental change rather than only static visual recognition.

\subsection*{Index Change Gate}

The index change gate filters mined events by checking whether the observed spectral change is strong enough to support a valid reasoning query. For each candidate event, the relevant index layers are computed or retrieved across the selected time window, and the magnitude of change is measured using predefined class thresholds. Samples are retained only when the change pattern is spatially coherent and semantically meaningful, such as a measurable decrease in vegetation, an increase in burn severity, or a detectable shift in built-up area. This step prevents weak or noisy temporal differences from entering the dataset and ensures that index-based questions have verifiable answers through tool execution.

\subsection*{Unified Data Representation}

After filtering, samples from the RGB-SAR, GIS, and index-based branches are converted into a unified data representation. Each record stores the image path or raster reference, GeoPackage information where applicable, object annotations, spatial metadata, candidate tool arguments, and task category. For GIS and index samples, the representation also includes area names, boundaries, coordinate reference information, POI layer names, index type, temporal range, and derived file references. This unified schema allows the question synthesizer and dataset constructor to operate over all modalities with a consistent interface, even though the original sources differ substantially in structure and annotation format.

\subsection*{Image and GeoPackage Information}

The image and GeoPackage information module stores the physical inputs required for tool execution. For image-based tasks, it records the image identifier, file path, image size, sensor type, and annotation references. For geospatial tasks, it stores GeoPackage paths, layer names, coordinate reference systems, boundary geometries, POI layers, and raster metadata. This information is passed to the tool orchestrator so that generated reasoning traces can refer to executable inputs rather than free-form descriptions. By keeping file references and spatial metadata explicit, the dataset supports deterministic replay and validation of tool-based trajectories.

\subsection*{Question Prompt Suite}

The question prompt suite contains task-specific templates used to synthesize natural-language queries. Each template corresponds to a reasoning category, such as object counting, object localization, region-attribute reasoning, proximity comparison, GIS distance computation, area estimation, index mapping, or temporal change analysis. The templates are designed to vary wording while preserving the required tool dependencies. For example, a GIS distance template specifies the target area, relevant POI classes, and distance relation, while an index-change template specifies the index type, region, and temporal range. This prompt suite encourages linguistic diversity while maintaining compatibility with the available annotations and tools.

\subsection*{Question Synthesizer}

The question synthesizer converts unified data records into natural-language queries and initial tool-grounded task specifications. It uses the question prompt suite together with the sample metadata to generate queries whose answers require one or more tool calls. For each generated query, the synthesizer also prepares the expected input references and candidate tool arguments. The objective is to produce questions that are natural to users but still grounded in the available imagery, annotations, GIS layers, or index products. Queries that contain mismatched objects, invalid regions, missing dates, or unsupported tool requirements are rejected or regenerated.

\subsection*{Conversational One-Shot Generator}

The conversational one-shot generator produces structured examples that guide the construction of reasoning trajectories. Each one-shot example demonstrates how an agent should decompose a query into thoughts, actions, observations, and a final answer. The generator is conditioned on the task category so that RGB-SAR, GIS, and index-based queries receive different reasoning patterns. For instance, an object-localization query may require detection followed by drawing a box, while a GIS query may require area-boundary extraction, POI-layer creation, distance computation, and map display. This component encourages consistent trajectory formatting across the corpus.

\subsection*{Dataset Prompt Suite}

The dataset prompt suite provides the instruction format used to generate complete training instances. It defines how the query, multimodal inputs, tool registry, examples, and output constraints are presented to the generator. The prompt suite enforces a structured reasoning format in which the agent alternates between thought, action, observation, and answer fields. It also constrains the generator to use valid tool names, valid argument names, and input references from the unified data record. This is important because the final dataset is intended not only for language supervision but also for executable tool-policy learning.

\subsection*{Dataset Constructor}

The dataset constructor combines the natural-language query, multimodal input references, unified metadata, and generated reasoning trace into a complete dataset entry. Each entry contains the user query, associated image or geospatial files, serialized tool calls, intermediate observations, and final answer. The constructor also records the task category, modality type, tool sequence, and reasoning length. This produces a standardized training or evaluation instance that can be consumed directly by the OpenEarthAgent training pipeline. By preserving intermediate steps, the dataset supports supervision over the complete reasoning process rather than only the final answer.

\subsection*{Validation Suite}

The validation suite performs automatic and manual checks before an instance is admitted into the final corpus. First, generated tool calls are parsed and checked for syntactic validity, including tool names, argument names, argument types, and file references. Second, the trajectory is replayed with the tool controller to verify that each action can be executed and that the returned observation is consistent with the next reasoning step. Third, spatial checks are applied to validate coordinates, geometries, POI layers, distances, areas, and raster-index outputs. Instances that fail validation are discarded or regenerated. For the held-out evaluation split, additional manual inspection is applied to remove unrealistic measurements, ambiguous queries, and cases where the answer depends on invalid metadata. The resulting samples form the final OpenEarthAgent dataset used for training and benchmarking.

\end{document}